\documentclass[journal]{IEEEtran}

\ifCLASSINFOpdf

\else

\fi

\usepackage{url}
\usepackage{amssymb}
\usepackage{amsthm}
\usepackage{amsmath}

\usepackage{epsfig}
\usepackage{amsmath,amssymb} % define this before the line numbering.
\usepackage{mathtools}
\usepackage{color}
\usepackage{algorithm}
\usepackage{cases}
\usepackage{multirow}
\usepackage{multicol}
\usepackage{algorithmicx}
\usepackage{graphicx}
\usepackage{makecell}
\begin{document}
\title{Learnable Subspace Clustering}
%\author{Jun~Li,~\IEEEmembership{Member,~IEEE,} Hongfu Liu,~\IEEEmembership{Member,~IEEE,} Zhiqiang Tao,~\IEEEmembership{Student Member,~IEEE,} \\
 %Handong Zhao~\IEEEmembership{Member,~IEEE,}~and~Yun~Fu,~\IEEEmembership{Fellow,~IEEE}

\author{Jun~Li,~\IEEEmembership{Member,~IEEE,} Hongfu Liu, Zhiqiang Tao, Handong Zhao~and~Yun~Fu,~\IEEEmembership{Fellow,~IEEE} % <-this % stops a space
\thanks{This work was partially done when Jun Li was PostDoc at Northeastern University.}
\thanks{Jun Li with the School of Computer Science and Engineering, Nanjing University of Science and Technology, Nanjing, China, 210094. Hongfu Liu is with Michtom School of Computer Science, Brandeis University, Waltham, MA 02453. Handong Zhao is a research scientist at Adobe Research, San Jose, CA, 95113. Zhiqiang Tao is with the Department of Electrical and Computer Engineering, College of Engineering, Northeastern University, Boston, MA, 02115. E-mail: junl.mldl@gmail.com; hongfuliu@brandeis.edu; hazhao@adobe.com; zqtao@ece.neu.edu.}
\thanks{Yun Fu is with the Department of Electrical and Computer Engineering, College of Engineering, and Khoury College of Computer Sciences, Northeastern University, Boston, MA 02115 USA (E-mail: yunfu@ece.neu.edu).}
%\thanks{Manuscript received March, 2019. }
}

\markboth{IEEE Transactions on Neural Networks and Learning Systems}% IEEE Transactions on Neural Networks and Learning Systems
{Shell \MakeLowercase{\textit{et al.}}: Bare Demo of IEEEtran.cls for Journals}

\maketitle

\begin{abstract}
This paper studies the large-scale subspace clustering (LSSC) problem with million data points. Many popular subspace clustering methods cannot directly handle the LSSC problem although they have been considered as state-of-the-art methods for small-scale data points. A basic reason is that these methods often choose all data points as a big dictionary to build huge coding models, which results in a high time and space complexity. In this paper, we develop a learnable subspace clustering paradigm to efficiently solve the LSSC problem. The key idea is to learn a parametric function to partition the high-dimensional subspaces into their underlying low-dimensional subspaces instead of the expensive costs of the classical coding models. Moreover, we propose a unified robust predictive coding machine (RPCM) to learn the parametric function, which can be solved by an alternating minimization algorithm. In addition, we provide a bounded contraction analysis of the parametric function. To the best of our knowledge, this paper is the first work to efficiently cluster millions of data points among the subspace clustering methods. Experiments on million-scale datasets verify that our paradigm outperforms the related state-of-the-art methods in both efficiency and effectiveness. Code is available at \emph{\url{https://github.com/junli2019/LeaSC}}
\end{abstract}


\IEEEpeerreviewmaketitle

\section{Introduction}
\label{introduction}
\IEEEPARstart{H}{igh}-dimensional big data are upsurgingly everywhere and are becoming more available and popular in computer vision and machine learning tasks. For example, millions of images \cite{ImageNet15}, and millions of videos with billions of frames \cite{Haija2016YouTube-8M} widely exist in Websites and YouTube, respectively. Unfortunately, the high-dimensionality of the big data usually leads to high time and space complexity of algorithms, and complex errors (e.g., noises, outliers, and corruptions) in the big data heavily hurt their performance \cite{Bellman1957dp}. However, the high-dimensional big data often lie in low-dimensional structures \cite{Maaten2009dr}. Therefore, finding low-dimensional structures in the big data becomes a fundamental problem to reduce the time and space complexity, cut down the effect of the complex errors, and furthermore improve the performance in learning and segmentation tasks.

Subspace clustering \cite{Vidalr2011} is one of the most common methods to robustly recover the low-dimensional representations of high-dimensional data since it has already provided theoretical guarantees \cite{Candes2011,Soltanolkotabi2014rsc} to successfully apply into numerous research areas, such as image segmentation \cite{Yang2008imgseg,Zhaohvs2015}, human motion segmentation \cite{Lis2015mseg}, image processing \cite{Ma2007seg,zhao2017aaai}, sequential data clustering \cite{Tierney2014seq}, medical imaging \cite{Ceting2014medicalseg}, and bioinformatics \cite{Wallace2015DNA}. Particularly, spectral-style methods (e.g., sparse subspace clustering (SSC) \cite{Elhamifar2013}, low-rank representation (LRR) \cite{Liugc2013,Zhangcq2015} and least squares regression (LSR) \cite{Lucy2012}) have attracted more and more attention in recent years due to the promising performance on small-scale datasets (e.g., Hopkins155 \cite{Tron2007Hopkins155} and Extended YaleB \cite{Georghiades2001eyaleb}). These methods are usually rooted in a \emph{self-expressiveness} (SE) property (i.e., each data point is represented by linearly combining other data points) \cite{Elhamifar2013}. Following this SE property, a classical paradigm is to choose all data points as a self-expressive dictionary in parsimonious coding models \cite{Elhamifar2013,Liugc2013}, solve the parsimonious coding models to obtain ideal sparse or low-rank representations (or codes), and use them to construct a similarity matrix for spectral clustering \cite{Ng2001sc}.  %times-series \cite{Vergara2013gas}
%\begin{quote}
%\textbf{Definition 1.1 (Large-scale Subspace Clustering).} \emph{Given a large number of data points drawn from a union of high-dimensional subspaces included errors (e.g., noise, outliers, and corruptions), segment the high-dimensional subspaces into their underlying low-dimensional subspaces and separate the including errors.}
%\end{quote}

\begin{figure}[t]
\vskip -0.0in
\begin{center}
\centerline{\includegraphics[width=0.9\columnwidth]{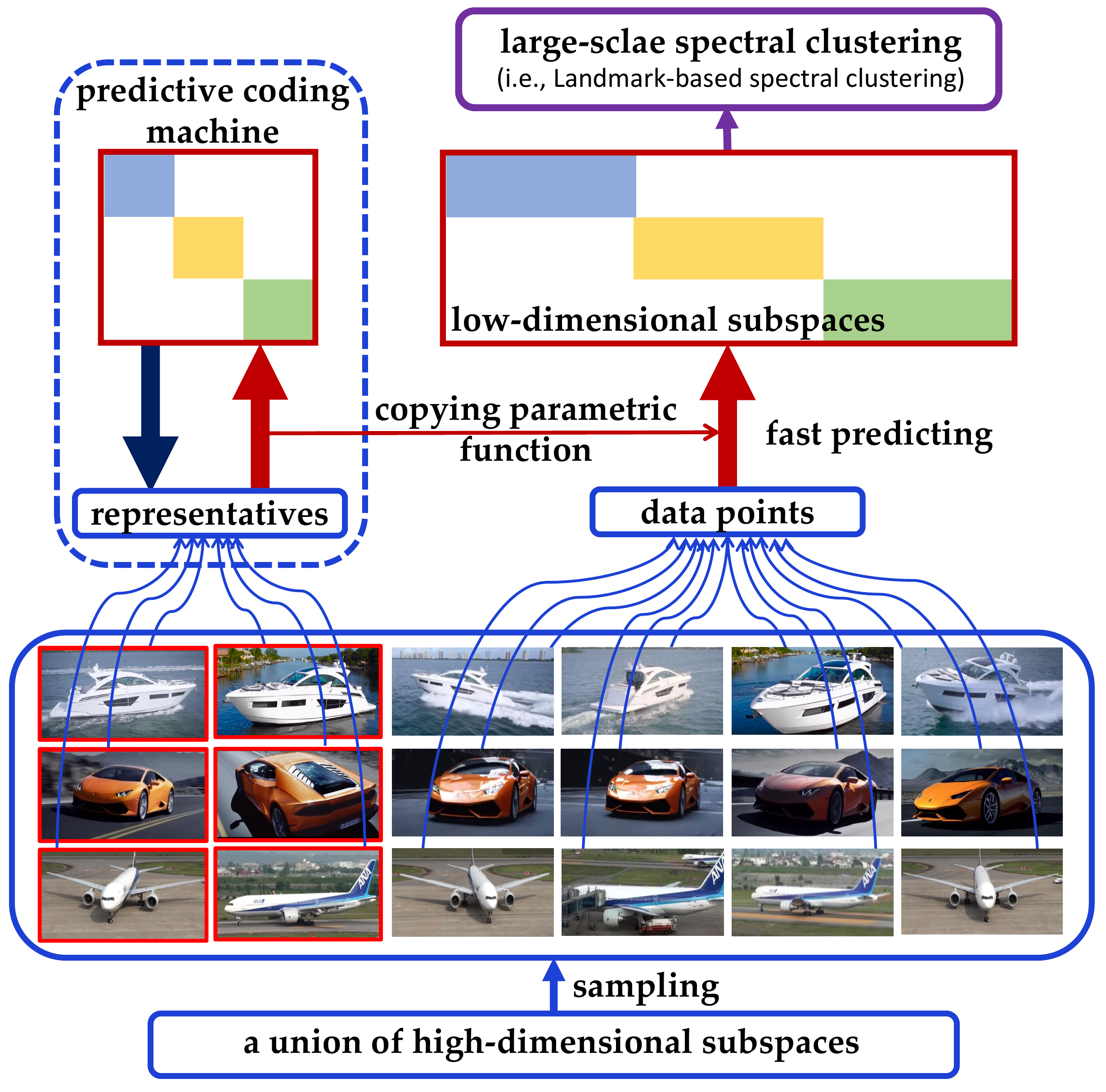}}
\vskip -0.1in
\caption{Illustration of learnable subspace clustering.}
\label{fig:LeaSCparadigm}
\end{center}
\vskip -0.4in
\end{figure}

%when the classical paradigm is applied into handle the high-dimensional big data (e.g., millions of images \cite{loosli-canu-bottou-2006} and videos \cite{Haija2016YouTube-8M}), thus
 %(similarities between data samples)
%the data-center subspace recovery. When facing the big data,
However, the classical paradigm restricts the subspace clustering methods to handle with the high-dimensional big data (e.g., millions of images \cite{loosli-canu-bottou-2006} and videos \cite{Haija2016YouTube-8M}). A basic reason is that the SE property leads to a big self-expressive dictionary, which is used to build huge parsimonious coding models. These coding models result in a major computational bottleneck since they are generally solved by iterating a lot of optimization operators, such as the shrinkage thresholding operator \cite{Becka2009ista}, and the singular value decomposition \cite{Golub1965svdpin}. Thus, it naturally arises a challenging subspace clustering problem with millions of data points.

%which learns a pattern function from the high-dimensional subspaces to the low-dimensional subspaces. ,
To address this problem, we develop a \emph{learnable subspace clustering} (LeaSC) paradigm (see Fig. \ref{fig:LeaSCparadigm}) to fast and robustly recover the low-dimensional structures from the high-dimensional big data. First, given a large set of high-dimensional data points, which are often drawn from a union of high-dimensional subspaces \cite{Elhamifar2013,Liugc2013}, we find some \emph{representatives} or \emph{exemplars} that can efficiently describe the drawn data points \cite{Elhamifar2016subset} to effectively reduce the size of the big self-expressive dictionary. Second, the representatives are used to learn a parametric function (e.g., neural networks) from the high-dimensional subspaces to the low-dimensional subspaces. Third, the rest data points are quickly mapped into the low-dimensional representations by the learned parametric function instead of the parsimonious coding models. Finally, we employ the landmark-based spectral clustering (LbSC) \cite{Caid2015} to cluster the representations.

In the developed LeaSC paradigm, we propose a novel model named \emph{Predictive Coding Machine} (PCM) to learn the parametric function. By finding the representatives as a small self-expressive dictionary, the basic procedure of PCM is to optimize the ideal regression, sparse, low-rank or elastic net codes by using a \emph{square of Frobenius norm}, \emph{$\ell_1$-norm}, \emph{nuclear norm} or \emph{Elastic net} regularization, and jointly train the parametric function to approximate the ideal codes. Moreover, since the data are often corrupted by noise (e.g., Gaussian noise, sparse outlying entries and missing entries) because of measurement/process noise in real-world applications \cite{Elhamifar2013,Liugc2013}, we practically extend PCM into a robust PCM (RPCM) by considering the data contaminated with the noises. Due to the nonlinearity of the parametric function, RPCM is non-convex. We solve them by proposing a quasi-convex optimization method, which alternates Alternating Direction Method of Multipliers (ADMM) \cite{Dengw2017pmbadmm,lu2018unified} and a gradient descent (GD) algorithm \cite{Lij2015sdsn,Lij2017SAinDNN}. By choosing a proper number of the representatives, we provide a theoretical condition to guarantee that the parametric function can well contract the high-dimensional data points to the ideal codes of the representatives. This further verifies that PCM and RPCM trained on the representatives can efficiently perform subspace clustering with millions of data points. Overall, our \textbf{main contributions} are as follows.
\begin{itemize}
\item We develop an effective paradigm, termed LeaSC, to deal with subspace clustering problem with millions of data points in a highly efficient way. Our goal is to learn the parametric function to partition the high-dimensional subspaces into their underlying low-dimensional subspaces instead of the expensive costs of the classical coding models. This parametric function leads to the linear complexity of LeaSC.
    %To the best of our knowledge, we are the first efficient one to cluster million data points in the subspace clustering methods. Our goal is to learn a parametric function from the high-dimensional subspaces to the low-dimensional subspaces.
\item We propose a predictive coding machine (PCM) model and its robust extension named RPCM to learn the parametric function in the LeaSC paradigm. Since RPCM is non-convex, we present a quasi-convex optimization method to solve it by alternating ADMM and GD.
\item We provide a bounded contraction analysis of the parametric function learned by our PCM model. It shows that the parametric function can well calculate contractive low-dimensional representations of high-dimensional data compared to the original subspace clustering models.
\end{itemize}

This paper is an important extension to our several previous conference works \cite{Lij2017frc,Lij2017plrsc,Lij2017sscl0,Lij2018pcm}. There are four major differences compared to the conference version. First, we develop a unified LeaSC paradigm to cluster millions of data points. Second, we propose a general RPCM with square of Frobenius norm \cite{Lij2017frc}, nuclear norm \cite{Lij2017plrsc}, $\ell_1$-norm, or Elastic net regularization. Third, we provide a theoretical condition to guarantee the contraction of the learned parametric function. Fourth, we add more experiments on million-scale datasets (i.e., MNIST8M \cite{loosli-canu-bottou-2006} and YouTube8M \cite{Haija2016YouTube-8M}) to show the effectiveness of LeaSC.

The rest of this paper is organized as follows. We review the related subspace clustering work in Section \ref{sec:relatedwork}. Our unified LeaSC paradigm and RPCM are developed in Section \ref{sec:LeaSCparadigm}. We propose a quasi-convex optimization method for RPCM in Section \ref{sec:ncoforRPCM}. We provide a contractive subspace recovery theory in Section \ref{sec:theoreticalanalysis}. Section \ref{sec:experiments} shows the experimental results. Finally, the conclusions are draw in Section \ref{sec:cons}.

\section{Related Work}
\label{sec:relatedwork}
%Many other subspace clustering methods, such as iterative, algebraic and statistical methods, can be found in \cite{Bradley2000kpc,Vidal2005gpca,Tipping1999ppca,Vidalr2011}.
In this section, we mainly review classical coding models, large-scale spectral clustering methods, and scalable coding methods as they have shown good performances on small-scale datasets. In addition, we discuss the direct encoding models to apply our LeaSC paradigm.

%They classically employ different coding models to compute a similarity matrix, and apply (large-scale) spectral clustering \cite{Ng2001sc} to this similarity matrix. Hence, we ,

\subsection{Classical Coding Models}
In general, rooting in the SE property \cite{Elhamifar2013}, the classical coding models employ different coding models to compute a similarity matrix, and apply (large-scale) spectral clustering \cite{Ng2001sc} to this similarity matrix. For example, SSC \cite{Elhamifar2013}, LRR \cite{Liugc2013}, and LSR \cite{Lucy2012} build the similarity matrix by using the coding models with $\ell_1$, nuclear norm and $\ell_2$ regularization, respectively. Based on the SE property, these coding models and their variants (e.g., \cite{Yuanxt2014,Zhangcq2015,Zhu2019mvsc,Yang2019sc}) usually choose all data points as a dictionary. When facing big datasets (e.g., million images), the number of autos (or bases) in the dictionary is very large. Clearly, this large dictionary will lead to over high-dimensional (million-by-million) similarity matrices for the spectral clustering in practice, and further result in that the coding models fail to run on a single machine if it has a limited resource. Moreover, they still suffer from a high time complexity as they are solved in iterative optimization manners \cite{Wangss2014,Pengx2016sc}. Hence, the parsimonious coding models take the high computational time and the large memory to build the similarity matrix. In addition, distributed LRR (DLRR) \cite{Talwalkara2013} and decentralized SSC (DSSC) \cite{Liu2016DRSC} partition a big dataset into many small ones, and use LRR or sparse coding to compute the low-rank or sparse representations. However, they still need a large amount of computer resources. Fortunately, we explore the LeaSC paradigm to learn a parametric function to quickly compute the representations for the similarity matrix.% by selecting some representatives.

\subsection{Large-scale Spectral Clustering}

General spectral clustering method \cite{Ng2001sc} needs to calculate the eigenvectors of a normalized Laplacian matrix formed from the similarity matrix, and apply $K$-means to cluster the eigenvectors. Unfortunately, calculating the eigenvectors is very expensive \cite{Caid2015}. Thus, it leads to a difficult problem to handle large-scale datasets. Hence, the spectral clustering method is extended into large-scale spectral clustering (LsSpC) to cluster the big data as the following two ways. First, the Nystr$\ddot{o}$m method is used to approximate the eigenvectors \cite{Fowlkesc2004} for reducing the computational cost, and the eigenvalue computations are paralleled in many subsystems \cite{Chenwy2011}. Second, a small number of samples is selected as landmark points to construct a sparse similarity matrix for, such as, $K$-means \cite{Yand2009}, out-of-sample \cite{Nief2011}, and random selection \cite{Caid2015}. They will lead to poor clustering results because an indivisible similarity matrix is often built by the original data with complex structure. By comparison, we can learn robust representations by using our RPCM model in the developed LeaSC paradigm.

\subsection{Scalable Coding Methods}

Scalable coding methods are also proposed to cluster big data as the following two strategies. First, scalable SSC \cite{Youc2016sssc,Youc2016ensc} is to solve a sequence of subproblems with a small sub-dictionary for decreasing the computational time. Second, sampling-clustering-classification (S-C-C) \cite{Pengx2016sc,Wangss2014} selects a small sub-dataset from the large dataset to preform subspace clustering by using SSC, LRR and LSR, and learns a simple linear classifier or collaborative representation based classifier (CRC) \cite{Zhangl2011crc} to obtain the final clustering results. Unfortunately, scalable SSC still needs more computational time, for example, it spends 1,000 seconds to handle 100,000 data samples \cite{Youc2016sssc}, and S-C-C will result in poor clustering results because the simple classifiers are not easy to identity the complex samples \cite{Bengio2013}. By comparison, we can fast calculate the robust representations to construct the similarity matrix, and perform the LsSpC (e.g., LbSC \cite{Caid2015}).

% \subsection{Deep subspace clustering methods}

\subsection{Direct Encoding Models}
In our LeaSC paradigm, we propose the RPCM model to learn the parametric function to directly encode the representations of data points for reducing the computational complexity. Actually, there are some direct encoding models, such as, auto-encoders (AE) \cite{Bourlard1988AE} sparse auto-encoders (SAE) \cite{Ranzato2007sp,Luow2017csae}, denoising auto-encoders (DAE) \cite{Vincent2008dae}, predictive sparse decomposition (PSD) \cite{Gregor2010fsc,Lij2017sscl0}, robust principal component analysis encoder (RPCAec) \cite{Sprechmann2015}, and latent LRR (latLRR) \cite{Liu2011}. In fact, both AE and SAE are not robust models and require the clean input data. Although DAE and PSD can learn robust codes, it still needs the clean input data. Unfortunately, the real data often have many noises \cite{Elhamifar2013,Liugc2013}. To deal with the noises, RPCAec \cite{Sprechmann2015} and latLRR \cite{Liu2011} could quickly calculate the robust low-rank codes by separating the noises. However, RPCAec \cite{Sprechmann2015} only handles with a single low-rank subspace, and latLRR \cite{Liu2011} only learns a linear predictor, which is difficult to capture the complex data structure \cite{Bengio2013,Lij2017SAinDNN}. By comparison, our RPCM does not require the clear data, and learns a nonlinear predictor to calculate the robust representations. More importantly, we will also apply these encoding models into the LeaSC paradigm to deal with large-scale subspace clustering.

%AE learns an encoder from input data to feature codes, and reconstructs the input data by training a decoder map from the codes, while SAE adds a sparse regularization into AE, and DAE reconstructs the clean input from a artificially corrupted input data. But, both AE and SAE are not robust models. Although DAE can learn robust codes, it needs the clean input data. Unfortunately, the real data often have many noises \cite{Elhamifar2013,Liugc2013}. Moreover, PSD only has been applied in classification tasks, not subspace clustering, and RPCAec \cite{Sprechmann2015} assumed that the underlying data structure was a single low-rank subspace. In addition, although latLRR \cite{Liu2011} could quickly calculate the low-rank codes of multiple subspaces data by a linear predictor, it essentially resolved the insufficient sampling data \cite{Liu2011}, and the linear predictor was difficult to capture the complex data structure \cite{Bengio2013,Lij2017SAinDNN}.

\section{Learnable Subspace Clustering}
\label{sec:LeaSCparadigm}
We develop an efficient \emph{Learnable Subspace Clustering} (LeaSC) paradigm to cluster a large collection of multi-subspaces data points in this section. Specifically, we first introduce a machine learning problem for subspace clustering. To solve this problem, we propose a unified predictive coding machine by learning a parametric function from the high-dimensional space into the low-dimensional space. Then the parametric function is used to fast recover subspace representations of data points. Finally, we perform the landmark-based spectral clustering (LbSC) \cite{Caid2015} on the subspace representations for final clustering results.

\textbf{Notations}: $\mathcal{X}$ is denoted as high-dimensional space. Random variables, ordinary vectors, matrices and matric blocks are respectively written in uppercase, lowercase, uppercase and blackboard bold, for example, $X$, $\textbf{x}$, $\textbf{X}$ and $\mathbb{X}\hspace{-0.05cm}=\hspace{-0.05cm}\left(\hspace{-0.1cm}\begin{array}{cccc}
\textbf{X}\\
\textbf{Y}\\
\end{array}\hspace{-0.1cm}\right)$. Expectation (discrete case, $p$ is probability mass) is denoted by $\mathrm{E}_{p(X)}(\mathcal{L}(X))=\sum_{\textbf{x}}p(X=\textbf{x})\mathcal{L}(\textbf{x})$. Given a matrix $\textbf{A}=[a_{ij}]\in \mathbb{R}^{d\times n}$, we denote the nuclear norm by $\|\textbf{A}\|_{\ast}=\sum_i\sigma_i(\textbf{A})$ (the sum of the singular values of $\textbf{A}$), $\ell_1$ norm by $\|\textbf{A}\|_1=\sum_{ij}|a_{ij}|$, $\ell_{2,1}$ norm by $\|\textbf{A}\|_{2,1}=\sum_{j}\|\textbf{a}_{:j}\|_2$, $\text{F}$-norm by $\|\textbf{A}\|_{\text{F}}=\footnotesize{\sqrt{\sum_{ij}(a_{ij})^2}}$, and square of $\text{F}$-norm by $\|\textbf{A}\|_{\text{F}}^2=\sum_{ij}(a_{ij})^2$. The number of $n$-combination from a set including $m$ elements is denoted by ${m \choose n}$.
% Let $\mathrm{I}_n$ denote the $n$-dimensional unit hypercube $[0,1]^n$, and the space of continuous functions on $\mathrm{I}_n$ is denoted by $\mathcal{C}(\mathrm{I}_n)$.

\subsection{Problem Statement}
\label{sec:LeaSCproblem} %Machine Learning Problem in Subspace Clustering

%The goal of our LSC is to learn a common pattern from the high-dimensional space $\mathcal{X}$ into the low-dimensional space $\mathcal{Z}$ for fast recovering subspace representations of data points.

%Specifically, we formulate a fast regression coding (FRC) model to learn the non-linear function, and then adopt an analytical method and a gradient descent algorithm to solve the FRC model. The landmark-based spectral clustering (LSC) \cite{Caid2015} is used to fast segment the codes of all data points computed by the function. Finally, we provide some theoretical guarantees to show that FRC recovers the subspace representations (codes) effectively.

%Let $\mathcal{Z}$ be the $d$-low-dimensional representation space where $d=\sum_{i=1}^kd_i$. Formally, we assume that the pairs $(X,Z)\in\mathcal{X}\times\mathcal{Z}$ are random variables according to an unknown joint distribution $P(X,Z)$ with corresponding marginals $P(X)$ and $P(Z)$, and the generative relationship from $Z$ to $X$ can be described as

%Problem 1.1 only briefly states what we will study.
In this subsection, an explicit machine learning problem is described for subspace clustering. Specifically, we consider a $d$-high-dimensional space $\mathcal{X}$, which is an unknown union of $s\geq 1$ linear or affine subspaces $\{\mathcal{X}_i\}_{i=1}^s$ of unknown low-dimensions $d_i=\text{dim}(\mathcal{X}_i)$ ($0<d_i<d$, $i=1,\cdots,s$). Formally, we assume that $X\in\mathcal{X}$ is a random variable according to an unknown distribution $p(X)$, and it can be factorized as
\begin{align}
\label{eq:variableX}
X=\textbf{B}Z=\left[
   \begin{array}{lll}
   \textbf{B}_1 \hspace{-0.2cm} & \cdots \hspace{-0.2cm}& \textbf{B}_k
   \end{array}
   \right] \left[
   \begin{array}{ll}
   \vspace{-0.1cm} Z_1 \\
   \hspace{0.15cm} \vdots  \\
   Z_k \\
   \end{array}
   \right],
\end{align}
where $\textbf{B}_i\in\mathbb{R}^{d\times d_i}$ is an unknown subspace base that linearly spans the $i$th data subspace, and $Z_i$ is a $d_i$-low-dimensional subspace representation variable. Our machine learning goal of the subspace clustering problem is to
\begin{center}
\emph{seek a parametric function $f(\bullet;\theta)$, which predicts $Z$ from \\ $X$, to partition the space $\mathcal{X}$ into the subspaces $\{\mathcal{X}_i\}_{i=1}^s$.}
\end{center}

%\emph{find the number of subspaces $k$, the subspace dimensions $\{d_i\}_{i=1}^k$ and the subspace bases $\{\textbf{B}_i\}_{i=1}^k$,}
To achieve this goal, according to the underlying distribution $p(X)$, an \emph{expected risk} of $f$ is defined as
\begin{align}
\label{eq:risk}
\min_{\theta}\ \mathrm{E}_{p(X)}\left[\mathcal{L}(Z,f(X;\theta))\right] \ \ \text{s.t.}\ \ X=\textbf{B}Z.
\end{align}
where $\mathcal{L}(Z,f(X;\theta))$ is a loss function between $Z$ and $f(X;\theta)$, for example, the squared error loss function $(Z-f(X;\theta))^2$. Due to the unknown $p(X)$, the expected risk cannot be directly measured in Eq. \eqref{eq:risk}. But we can consider an empirical distribution $p^0(X)$ defined by $N$ data points $\textbf{X}=[\textbf{x}_i]_{i=1}^n \in{\rm I\!R}^{d\times n}$ that we suppose to be an i.i.d. sample from the distribution $p(X)$. The corresponding \emph{empirical risk}\footnote{Note that the empirical risk can be used instead as an unbiased estimate (i.e., replacing $\mathrm{E}_{p(X)}$ by $\mathrm{E}_{p^0(X)}$) \cite{Vincent2010}. In essence, there is a difference between the empirical risk and the expected risk \cite{Buhmann1999ERA}, and this difference is not considered here since we focus on the empirical risk.} average over the $p^0(X)$ can be described as
\begin{align}
\label{eq:empiricalrisk}
\min_{\theta}\ \mathrm{E}_{p^0(X)}\left[\mathcal{L}(Z,f(X;\theta))\right] \ \ \text{s.t.}\ \ X=\textbf{B}Z.
\end{align}
After learning the parametric function $f(\bullet;\theta)$, it is used to fast map the space $\mathcal{X}$ into low-dimensional representation spaces, which are easily divided into the subspaces $\{\mathcal{X}_i\}_{i=1}^s$. We will propose a robust and effective model to better calculate the \emph{empirical risk} in the next subsection.

\emph{\textbf{Remark 1.}} Our machine learning problem is different from the conventional subspace clustering problem \cite{Vidalr2011}. The latter often considers the segmentation of the given (training) data points $\textbf{X}$ drawn from the space $\mathcal{X}$ by using expensive sparse/low-rank optimizations. The former is to learn a parametric function for partitioning the space $\mathcal{X}$ into the subspaces $\{\mathcal{X}_i\}_{i=1}^s$. Its procedure is to draw a dataset from the space $\mathcal{X}$ or select a small set from the given $\textbf{X}$, and train the parametric function. This leads to a clear benefit that the parametric function can quickly project new (big) data points into low-dimensional representations to group into their own subspaces.
%\textbf{Remark 1.} The conventional subspace clustering problems often use the relationship in Eq. \eqref{eq:variableX} to build the coding models by adding the sparse/low-rank/regression regularization, such as SSC \cite{Elhamifar2013}, LRR \cite{Liugc2013} and LSR \cite{Lucy2012}. However, it is well-known that there is an expensive cost to iteratively optimize the subspace representation of each data point in these coding models. To avoid the expensive cost, LSC is to train the non-iterative mapping $Z=f(X;\theta)$ to quickly calculate the the ideal subspace representation of any data point. Thus, LSC can handle up to million data points.
\subsection{Predictive Coding Machine}
\label{sec:RPCM}
% To

We propose a novel predictive coding model to evaluate the empirical risk in Eq. \eqref{eq:empiricalrisk} by using the data $\textbf{X}$. Due to unknown the subspace bases $\textbf{B}$, we consider the self-expressiveness property \cite{Elhamifar2013}. This property implies that the data $\textbf{X}$ can be regarded as a self-expressive dictionary instead of the bases $\textbf{B}$, that is, $\textbf{X}=\textbf{X}\textbf{Z}$ with $\text{diag}(\textbf{Z})=0$, which overcomes the trivial solution of reconstructing a data point using itself. By considering the squared error loss function, thus, the Eq. \eqref{eq:empiricalrisk}  can be rewritten as:
\begin{align}
\label{eq:pcmmodel}
\min_{\textbf{Z},\theta}\  \|\textbf{Z}-f(\textbf{X};\theta)\|_{\text{F}}^2,\   \text{s.t.}\  \textbf{X}=\textbf{X}\textbf{Z}, \ \text{diag}(\textbf{Z})=0,
\end{align}
where $f(\textbf{X};\theta)$ is considered as a fully-connected network:
\begin{align}
\label{eq:deepfcnn}
f(\textbf{X};\theta)=g(\textbf{W}^l\cdots g(\textbf{W}^2\textbf{X})),
\end{align}
where $l$ is the number of layers, $g$ is an activation function (e.g., \emph{tanh} $g(a)=\frac{e^a-e^{-a}}{e^a+e^{-a}}$, \emph{sigmoid} $g(a)=\frac{1}{1+e^{-a}}$, \emph{ReLU} $max(0,a)$, and rectifier piecewise linear units \cite{Lij2017SAinDNN}), a learnable parameter set $\theta$ is considered as $\{\textbf{W}^2,\cdots,\textbf{W}^l\}\in \widetilde{{\rm I\!R}}=\{{\rm I\!R}^{\iota_{2}\times \iota_1},\cdots,{\rm I\!R}^{\iota_{l}\times \iota_{l-1}}\}$, and $\iota_i$ is the number of hidden units in the $i$-th layer ($\iota_1=d$ and $\iota_l=n$). For example, a fully-connected network with $l=4$ layers is $f(\textbf{X};\theta)=g(\textbf{W}^4g(\textbf{W}^3g(\textbf{W}^2\textbf{X})))$.

\textbf{Regularization.} However, the overcomplete dictionary $\textbf{X}$ (i.e., $n>d$) in Eq. \eqref{eq:pcmmodel} leads to the infinite number of subspace representations $\textbf{Z}$. In such case, the infinite $\textbf{Z}$ will result in trouble training the neural network in Eq. \eqref{eq:deepfcnn}. To overcome this, the regression/sparse/low-rank regularization is naturally used to restrain the subspace representations $\textbf{Z}$ (e.g., $\|\textbf{Z}\|_{\text{F}}^2$, $\|\textbf{Z}\|_{1}$, $\|\textbf{Z}\|_{\ast}$, and mixing $\|\textbf{Z}\|_{\text{F}}^2$ with $\|\textbf{Z}\|_{1}$. Particularly, $\|\textbf{Z}\|_{\text{F}}^2$ can restrict its solution space \cite{Lucy2012}; $\|\textbf{Z}\|_{1}$ is employed to obtain subspace-sparse representations since the number of each subspace base is greater than its intrinsic dimension \cite{Elhamifar2013}; $\|\textbf{Z}\|_{\ast}$ is used to recover the underlying row subspaces of data points as they are approximately drawn from a union of low-rank subspaces \cite{Liugc2013}.

%and mixing $\|\textbf{Z}\|_{1}$ with $\|\textbf{Z}\|_{\ast}$)
%\textbf{Contraction.} Moreover, to extract robust subspace representations $f(\textbf{X};\theta)$, we reduce its sensitivity to the input data points $\textbf{X}$ by penalizing a square of Frobenius norm of the Jacobian term $J_f(\textbf{X};\theta)$, which is defined as:
%\begin{align}
%\label{eq:jacfcnn}
%\|J_f(\textbf{X};\theta)\|_{\text{F}}^2=\|\frac{\partial f(\textbf{X};\theta)}{\partial \textbf{X}}\|_{\text{F}}^2.
%\end{align}
%Penalizing this term drives $f(\textbf{X};\theta)$ to the subspace to be contractive in the neighborhood of $\textbf{X}$. The low valued $\|J_f(\textbf{X};\theta)\|_{\text{F}}^2$ directly implies an invariance of the subspace representations for small variations of $\textbf{X}$. Thus, this invariance will be further used to exploit robust subspace representations in section \ref{sec:theoreticalanalysis}.

\textbf{Our PCM model.} By adding the above mentioned regularization and contraction, we propose a \emph{predictive coding machine} (PCM) to specify the empirical risk in Eq. \eqref{eq:pcmmodel} as
\begin{align}
\label{eq:RPCMmodel-1}
\min_{\textbf{Z},\theta}\ & \|\textbf{Z}-f(\textbf{X};\theta)\|_{\text{F}}^2+ \overline{\alpha} R_1(\textbf{Z})+ \alpha R_2(\textbf{Z}), \\
 \text{s.t.}\ & \textbf{X}=\textbf{X}\textbf{Z}, \ \text{diag}(\textbf{Z})=0, \nonumber
\end{align}
where $R_1(\textbf{Z})$ and $R_2(\textbf{Z})$ are the unified regularization forms of the representations $\textbf{Z}$, $\overline{\alpha}$  and $\alpha$ are regularization parameters. (In this paper, $\overline{\alpha}=1$ for simplicity). To understand the PCM model in Eq. \eqref{eq:RPCMmodel-1}, $\alpha=0$ for RPCM$_{\text{F}^2/\ell_1/\ast}$ except RPCM$_{\ell_1\hspace{-0.05cm}+\hspace{-0.01cm}\text{F}^2}$. In addition, we consider four regularization choices of $\textbf{Z}$ shown in Table \ref{tab:fivenorms}, while mixing low-rank and sparse regularization is not studied as it cost much more time to optimize the codes. Note that since the  fully-connected network is used to compute the current $\textbf{Z}$ to update other variables in section \ref{sec:ncoforRPCM}, we do not use an additional parameter to balance the parameter-learnable term and the regularization terms in the PCM model.

%$\alpha=(\alpha_1,\alpha_2)$ when $R(\textbf{Z})$ is a multiple regularization (i.e., $\left(\|\textbf{Z}\|_{1},\|\textbf{Z}\|_{\text{F}}^2\right)^{\text{T}}$).
% $R(\textbf{Z},\textbf{E})=\alpha_1\|\textbf{Z}\|_{\text{F}}^2+\alpha_2\|\textbf{Z}\|_{1}+\alpha_3\|\textbf{Z}\|_{\ast}+\alpha_4\|\textbf{E}\|_{2,1}$, where $\alpha_i(1\leq i\leq 4)$ are the regularization parameters. %\\ \alpha_1\|\textbf{Z}\|_{\text{F}}^2+ \alpha_2\|\textbf{Z}\|_{1}+\alpha_3\|\textbf{Z}\|_{\ast},
% \multicolumn{3}{|l|}{}

\textbf{Robust PCM (RPCM).} It is a practical extension of PCM.
In real-world applications, the data $\textbf{X}$ are often corrupted by noise (e.g., gaussian noise, sparse outlying entries and missing entries) because of measurement/process noise \cite{Elhamifar2013,Liugc2013}. By considering the data $\textbf{X}$ contaminated with the noises $\textbf{E}$, it holds that $\textbf{X}=\textbf{X}\textbf{Z}+\textbf{E}$ with $\text{diag}(\textbf{Z})=0$. To deal with the different noises, we follow the SSC and LRR models \cite{Liugc2013,Elhamifar2013} and add a $\ell_{2}$ or $\ell_{1}$ or $\ell_{2,1}$-norm of the noises $\textbf{E}$ into the empirical risk in Eq. \eqref{eq:RPCMmodel-1} to separate these noises. By separating the noises, our RPCM in Eq. \eqref{eq:RPCMmodel-1} is rewritten as
\begin{align}
\label{eq:RPCMmodel}
\min_{\textbf{Z},\textbf{E},\theta}\ & \|\textbf{Z}-f(\textbf{X};\theta)\|_{\text{F}}^2+ \overline{\alpha} R_1(\textbf{Z})+ \alpha R_2(\textbf{Z})+ \beta R_3(\textbf{E}), \\
 \text{s.t.}\ & \textbf{X}=\textbf{X}\textbf{Z}+\textbf{E}, \ \text{diag}(\textbf{Z})=0, \nonumber
\end{align}
where $R_3(\textbf{E})$ are the unified regularization forms of the noises $\textbf{E}$ in Table \ref{tab:fivenorms} and $\beta$ is a regularization parameter for the noises.

\begin{table}[t]
\linespread{1}
\vskip -0.0in
\setlength{\tabcolsep}{1.3pt}
\renewcommand{\arraystretch}{1.3}
\caption{Four variants of RPCM with different regularization choices of $\textbf{Z}$ and $\textbf{E}$. }
\vskip -0.15in
\label{tab:fivenorms}
\begin{center}
%\scriptsize \footnotesize,
\begin{tabular}{c|c|c|c|c|c|c}
\Xhline{1.2pt}
variants & norms             & $R_1(\textbf{Z})$     & $R_2(\textbf{Z})$      & $R_3(\textbf{E})$                   & constraint       &  learnable \\
\hline
RPCM$_{\ell_1\hspace{-0.05cm}+\hspace{-0.01cm}\text{F}^2}$ & elastic-net  &  $\|\textbf{Z}\|_{1}$  & $\|\textbf{Z}\|_{\text{F}}^2$   & $\|\textbf{E}\|_{2,1}$   & $\text{diag}(\textbf{Z})\hspace{-0.1cm}=\hspace{-0.1cm}0$ & \multirow{4}{*}{$f(\bullet;\theta)$}   \\
RPCM$_{\ell_1}$ & $\ell_1$          & $\|\textbf{Z}\|_{1}$   & $-$      & $\|\textbf{E}\|_{2,1}$        &$\text{diag}(\textbf{Z})\hspace{-0.1cm}=\hspace{-0.1cm}0$ &     \\
RPCM$_{\ast}$ & nuclear           & $\|\textbf{Z}\|_{\ast}$   & $-$   & $\|\textbf{E}\|_{2,1}$      & $-$ & \\
RPCM$_{\text{F}^2}$ & square of $\text{F}$     & $\|\textbf{Z}\|_{\text{F}}^2$  & $-$ &$\|\textbf{E}\|_{\text{F}}^2$&$\text{diag}(\textbf{Z})\hspace{-0.1cm}=\hspace{-0.1cm}0$ &\\
%RPCM$_{\ell_1\hspace{-0.05cm}+\hspace{-0.05cm}\ast}$ & $\ell_1$+nuclear  & $\left(\|\textbf{Z}\|_{1},\|\textbf{Z}\|_{\ast}\right)^{\text{T}}$         & $\|\textbf{E}\|_{1}$ & $\text{diag}(\textbf{Z})\hspace{-0.1cm}=\hspace{-0.1cm}0$ &   \\
\Xhline{1.2pt}
\end{tabular}  %}
\end{center}
\vskip -0.2in
\end{table}

Ideally, the parameter solution $\theta$ in the RPCM model and its practical extension stores the pattern to project the high-dimensional space $\mathcal{X}$ into its low-dimensional space. Thus, the fully-connected network $f(\bullet;\theta)$ is used to fast build a similarity graph and quickly infer the clustering of huge data points sampled from $\mathcal{X}$ using the LbSC \cite{Caid2015} in next subsection. In Section \ref{sec:ncoforRPCM}, we propose a quasi-convex optimization and establish the convergence guarantee conditions for the problems in Eqs. \eqref{eq:RPCMmodel-1} and \eqref{eq:RPCMmodel}. In Section \ref{sec:theoreticalanalysis}, we study robust recovery conditions for the fully-connected network, which can calculate robust subspace representation of each data point compared to the parsimonious coding models in the popular subspace clustering methods (i.e., LSR \cite{Lucy2012}, SSC \cite{Elhamifar2013}, LRR \cite{Liugc2013,Zhangcq2015}, and ENSC \cite{Panagakis2014ENSC}). Moreover, by incorporating the corruption model of data into the Eq. \eqref{eq:RPCMmodel-1}, we can handle with clustering of corrupted data. To better understand the RPCM model, we have the following two remarks:

\emph{\textbf{Remark 2.}}
RPCM is a unified model with different regularization choices in Table \ref{tab:fivenorms}. Its five variants (i.e., RPCM$_{\text{F}^2}$, RPCM$_{\ell_1}$, RPCM$_{\ast}$, and RPCM$_{\ell_1\hspace{-0.05cm}+\hspace{-0.01cm}\text{F}^2}$) are built by extending the popular subspace clustering methods into a machine learning framework. Compared to the predictive $\ell_1$-sparse decomposition (PSD) \cite{Kavukcuoglu2008,Gregor2010fsc}, RPCM$_{\ell_1}$ is a robust model by separating noises and penalizing the sensitivity to the data points. Compared to the predictive non-negative matrix factorization (PNMF) \cite{Sprechmann2015}, RPCM$_{\ast}$ can deal with mutiple subspaces. In addition, our SSC-LOMP \cite{Lij2017sscl0} is also an extension of SSC-OMP \cite{Youc2016sssc} by using $\ell_0$-norm. In fact, both RPCM$_{\text{F}^2}$ \cite{Lij2017frc} and RPCM$_{\ast}$ \cite{Lij2017plrsc} are our previous works.

%and contractive auto-encoders (CAE) \cite{Rifai2011cae} Third, RPCM is also similar to CAE as they penalize the sensitivity to the data points. %In addition, RPCM is more robust than our PCM model \cite{Lij2018pcm}, and PCM is mainly applied into compressed sensing and image denoising, not subspace clustering.
\emph{\textbf{Remark 3.}} Compared to DAE \cite{Vincent2010}, RPCM can learn more robust subspace representations than DAE. First, RPCM is to separate the sparse outliers $\textbf{E}$ of the data points $\textbf{X}$ for learning their robust representations, while DAE requires the clear $\textbf{X}$ as it is trained to reconstruct clean ``repaired" $\textbf{X}$ from a corrupted version of them. Second, RPCM is similar to DAE since $\textbf{X}$ with sparse outliers is considered as a corrupted version to recover clean $\textbf{X}-\textbf{E}$.
\subsection{Clustering Using the Parametric Function}
%This implies that  However, the requirement is over strict as the data points often contain sparse outliers. Fortunately, . Moreover, corrupts the clear training data points and reconstruct them by training a
Given data points $\textbf{Y}\in{\rm I\!R}^{d\times m}$ (i.e., $m>1,000,000$), which are i.i.d. sampled from the distribution $p(X)$ ($X\in\mathcal{X}$), it is difficult to cluster such huge $\textbf{Y}$. To handle this problem, we present an LeaSC procedure by using RPCM.

% and obtain its representation matrix $\textbf{Z}^{\circ}$
First, we find $n$ \emph{representatives} $\textbf{X}\in{\rm I\!R}^{d\times n} (n<m)$ in the huge data $\textbf{Y}$, as $\textbf{X}$ can preserve the characteristics of $\textbf{Y}$ \cite{Elhamifar2016subset}. Moreover, $\textbf{X}$ is well chosen from $\textbf{Y}$ by employing $K$-means or dissimilarity-based sparse subset selection (DS3) algorithms, which can ensure a suitable Euclidean distance between the representatives and the data points in each subspace \cite{Elhamifar2016subset}. In practice, we can fast choose $\textbf{X}$ by using random selection.
Second, the selected representatives $\textbf{X}$ are used to efficiently train the parametric function $f(\bullet;\theta)$ by solving the proposed RPCM in Eq. \eqref{eq:RPCMmodel}. Then the representations $\textbf{Z}^{f_{\textbf{Y}}}\in{\rm I\!R}^{n\times m}$ of $\textbf{Y}$ are fast computed by the parametric function, that is, $\textbf{Z}^{f_{\textbf{Y}}}=f(\textbf{Y};\theta)$.
Third, the representations $\textbf{Z}^{f_{\textbf{Y}}}$ are applied to compute the similarity matrix $\mathcal{W}$ as:
\begin{align}
\label{eq:affinitymatrix}
\mathcal{W}=(\widetilde{\textbf{Z}}^{f_{\textbf{Y}}})^{\text{T}}\widetilde{\textbf{Z}}^{f_{\textbf{Y}}},
\end{align}
where $\widetilde{\textbf{Z}}^{f_{\textbf{Y}}}=\textbf{D}^{-\frac{1}{2}}|\textbf{Z}^{f_{\textbf{Y}}}|$, and $\textbf{D}$ is an $n\times n$ diagonal matrix with $\textbf{D}_{ii}=\sum_{j}|\textbf{Z}^{f_{\textbf{Y}}}_{ij}|$. Following the LbSC \cite{Caid2015}, we build an inexpensive method to calculate the eigenvectors of the similarity matrix $\mathcal{W}$ as follows:
\begin{align}
\label{eq:eigenvectors}
\textbf{V}^{\text{T}}=\Sigma^{-1}\textbf{U}^{\text{T}}\widetilde{\textbf{Z}}^{f_{\textbf{Y}}},
\end{align}
where $\Sigma\hspace{-0.08cm}=\hspace{-0.08cm}\text{diag}(\sigma_1,\cdots,\sigma_k)$ ($\sigma_1\hspace{-0.08cm}\geq\hspace{-0.08cm}\cdots\hspace{-0.08cm}\geq\hspace{-0.08cm}\sigma_k\hspace{-0.08cm}\geq\hspace{-0.08cm} 0$) are the positive square root of first $k$ singular values of matrix $\widetilde{\textbf{Z}}^{f_{\textbf{Y}}}(\widetilde{\textbf{Z}}^{f_{\textbf{Y}}})^{\text{T}}$, $\textbf{V}=(\textbf{v}_1,\cdots,\textbf{v}_k)$ are the first $k$ eigenvectors of $\widetilde{\textbf{Z}}^{f_{\textbf{Y}}}(\widetilde{\textbf{Z}}^{f_{\textbf{Y}}})^{\text{T}}$, $k$ is the number of clusters, and $\widetilde{\textbf{Z}}^{f_{\textbf{Y}}}$ is defined in Eq. \eqref{eq:affinitymatrix}.
Finally, \emph{$K$-means} is used to cluster rows of $\textbf{V}$ for segmenting the data points $\textbf{Y}$ into their corresponding subspaces.

%$\overline{\textbf{Z}}=\textbf{U}\Sigma\textbf{V}^{\text{T}}$ is the SVD of $\overline{\textbf{Z}}$, $\Sigma=\text{diag}(\sigma_1,\cdots,\sigma_N)$ ($\sigma_1\geq\cdots\geq\sigma_N\geq 0$) are the singular values of $\overline{\textbf{Z}}$, $\textbf{U}=(\textbf{u}_1,\cdots,\textbf{u}_N)\in{\rm I\!R}^{N\times N}$ are the left singular vectors, and $\textbf{V}=(\textbf{v}_1,\cdots,\textbf{v}_N)\in{\rm I\!R}^{M\times N}$ are right singular vectors.

% \emph{representatives-expressiveness property}  \emph{all representatives can be efficiently reconstructed by linear combinations of representatives in a union of subspaces.}
% Compared to the self-expressiveness property \cite{Elhamifar2013}, there is an important difference that it results in a big $D\times N$ self-expressive dictionary $\textbf{X}$ and over-high dimensional $N\times N$ subspace representations, while the introduced property leads to the small $D\times M$ dictionary $\textbf{Y}$ and low-dimensional $M\times N$ representations $\textbf{Z}$.

Overall, our LeaSC procedure is summarized in \textbf{Algorithm \ref{alg:learnsc}}. Note that LeaSC can efficiently handle up to million-scale datasets because RPCM learns a parametric function to fast compute the representations of data points and the eigenvectors of the affinity matrix can be quickly calculated by Eq. \eqref{eq:eigenvectors}. Compared to the computational complexities $\mathcal{O}(m^2)$ or $\mathcal{O}(m^3)$ of the conventional subspace clustering methods \cite{Elhamifar2013,Liugc2013}, the complexity of LeaSC is \emph{linear} in terms of $m$.

In addition, since the parameter $n$ is important to randomly select $\textbf{X}$ which simultaneously includes all the bases, we give a theoretical rule for setting $n$.

\textbf{Proposition 1.} (The proof is provided in the appendix A.) \emph{Corresponding to} $\widehat{n}$ \emph{bases in a union of subspaces,} $\textbf{X}=\{\textbf{X}_i\}_{i=1}^{\widehat{n}}$ \emph{is randomly selected} \emph{from} $\textbf{Y}=\{\textbf{Y}_i\}_{i=1}^{\widehat{n}}$\emph{, where} $\sum_{i=1}^{\widehat{n}}n_i=n$, $\sum_{i=1}^{\widehat{n}}m_i=m$, $n_i$ \emph{and} $m_i$ \emph{are the number of} $\textbf{X}_i$ \emph{and} $\textbf{Y}_i$. \emph{There is a probability} $P$ \emph{for} $\textbf{X}$ \emph{including at least one data point in every basis,}
\begin{align}
\label{eq:probselects}
P=\frac{\sum_{(n_1,\cdots,n_{\widehat{n}})\in\mathbf{S}}\left[\prod_{i=1}^{\widehat{n}} {m_i \choose n_i}\right]}{{m \choose n}},
\end{align}
\emph{where} $\mathbf{S}\hspace{-0.1cm}=\hspace{-0.1cm}\{(n_1,\cdots,n_{\widehat{n}})|\sum_{i=1}^{\widehat{n}}n_i\hspace{-0.1cm}=\hspace{-0.1cm}n \& \{1 \leq n_i\leq m_i\}_{i=1}^{\widehat{n}}\}$ \emph{is a set of all events (i.e., each} $\textbf{X}_i$ \emph{has at least one data point in} $\textbf{Y}_i$\emph{),} ${m \choose n}$ \emph{and} ${m_i \choose n_i}$ \emph{are the} $n$/$n_i$-combination \emph{of} $\textbf{Y}$ \emph{and} $\textbf{Y}_{i}$.

\begin{table*}[t]
\linespread{1}
\vskip -0.0in
\caption{Unified matrix blocks in RPCM. Note $\alpha=0$ for RPCM$_{\text{F}^2/\ell_1/\ast}$ except RPCM$_{\ell_1\hspace{-0.05cm}+\hspace{-0.01cm}\text{F}^2}$, and $\overline{\alpha}=1$ (It also can be tuned).}
\vskip -0.2in
\label{tab:unifiedweights}
\begin{center}
\begin{tabular}{c|c|c|c|c|c|c|c|c|c}
\Xhline{1.2pt}
$\mathbb{X}$ &  $\mathbb{A}_1$   &   $\mathbb{Z}$  & $\mathbb{A}_2$   &  $\mathbb{E}$ & $R(\mathbb{Z})$ & $R(\mathbb{E})$ & $\mathbb{Q}$ &$\widehat{\alpha}$ & $\widehat{\beta}$\\
 \hline
 $\left(\hspace{-0.1cm}\begin{array}{cccc}
0\\
\textbf{X}\\
\end{array}\hspace{-0.1cm}\right)$& $\left(\hspace{-0.2cm}\begin{array}{cccc}
\textbf{I}_N  \hspace{-0.2cm}& -\textbf{I}_N \\
0 \hspace{-0.2cm} & \textbf{X}\\
\end{array}\hspace{-0.2cm}\right)$ & $\left(\hspace{-0.1cm}\begin{array}{cccc}
\textbf{J}\\
\textbf{Z}\\
\end{array}\hspace{-0.1cm}\right)$ & $\left(\hspace{-0.2cm}\begin{array}{cccc}
\textbf{I}_N  \hspace{-0.2cm}& 0 \\
0 \hspace{-0.2cm} & \textbf{I}_M\\
\end{array}\hspace{-0.2cm}\right)$ & $\left(\hspace{-0.1cm}\begin{array}{cccc}
0\\
\textbf{E}\\
\end{array}\hspace{-0.1cm}\right)$ & $\left(\hspace{-0.1cm}\begin{array}{cccc}
R_1(\textbf{J})\\
R_2(\textbf{Z})\\
\end{array}\hspace{-0.1cm}\right)$ & $\left(\hspace{-0.1cm}\begin{array}{cccc}
0\\
R_3(\textbf{E})\\
\end{array}\hspace{-0.1cm}\right)$ & $\left(\hspace{-0.1cm}\begin{array}{cccc}
\textbf{Q}_1\\
\textbf{Q}_2\\
\end{array}\hspace{-0.1cm}\right)$ & $\left(\hspace{-0.1cm}\begin{array}{cccc}
 \overline{\alpha} \\
\alpha\\
\end{array}\hspace{-0.1cm}\right)$ & $\left(\hspace{-0.1cm}\begin{array}{cccc}
0\\
\beta\\
\end{array}\hspace{-0.1cm}\right)$ \\
\Xhline{1.2pt}
\end{tabular}
\end{center}
\vskip -0.2in
\end{table*}

This proposition shows that there are three cases to compute the probability: 1) $P=0$ if $0<n<\widehat{n}$; 2) $0<P<1$ if $\widehat{n}\hspace{-0.1cm}\leq \hspace{-0.1cm}n\hspace{-0.1cm}\leq\hspace{-0.1cm} \max_{i}\{m-m_i\}$; 3) $P=1$ if $\max_{i}\{m-m_i\}< n\leq m$.
Following the case 2), we have a empirical rule that given a large number $m$ ($m_i=m_j,i\neq j$), there is a suitable $n$ (e.g., $5\widehat{n}\leq n\leq \frac{(\widehat{n}-1)m}{\widehat{n}}$) to obtain a high probability. For example, if $m=100$, $\widehat{n}=2$, $n=3.5\widehat{n}=7$, $m_1=m_2=50$ and $\mathbf{S}=\{(1,6),(2,5),(3,4),(4,3),(5,2),(6,1)\}$, then $P=0.9875$. When the probability $P$ will drop with the increasing $m$, we also can increase the parameter $n$ to improve $P$.
\begin{algorithm}[tb]
\caption{Learnable Subspace Clustering via RPCM.}
\begin{algorithmic}[1]
\State {\bfseries Initialize:} large data $\textbf{Y}$.
\State \hspace{-0.11cm} Choose representatives $\textbf{X}$ from $\textbf{Y}$ using DS3 \cite{Elhamifar2016subset} or random selection,
\State \hspace{-0.11cm}  Train a parametric function $f(\bullet;\theta)$ by RPCM and $\textbf{X}$,
\State \hspace{-0.11cm}  Fast calculate all data representations by $\textbf{Z}^{f_{\textbf{Y}}}=f(\textbf{Y};\theta)$,
\State \hspace{-0.11cm}  Quickly compute the eigenvectors $\textbf{V}$ by $\textbf{Z}^{f_{\textbf{Y}}}$ and Eq. \eqref{eq:eigenvectors},
\State \hspace{-0.11cm}  Perform \emph{$K$-means} on $\textbf{V}$ to segment $\textbf{Y}$ into $k$ subspaces.
\end{algorithmic}
\label{alg:learnsc}
\vskip -0.0in
\end{algorithm}
\hspace{-1cm}
\section{Quasi-convex Optimization for RPCM}
\label{sec:ncoforRPCM}
RPCM in Eq. \eqref{eq:RPCMmodel} is a non-convex and non-smooth problem because of the nonlinear parametric function. We present an effective quasi-convex optimization strategy to alternate optimizing codes and training the nonlinear parametric function. Moreover, we discuss the convergence of the quasi-convex optimization, and its computational complexity.
%  \multicolumn{5}{|c|}{RPCM$_{\text{F}^2}$, RPCM$_{\ell_1}$, RPCM$_{\ast}$} \\

%. We consider Gauss-Seidel ADMM \cite{lu2018unified} By introducing an auxiliary variable $\textbf{J}$,
\subsection{Alternating ADMM and GD for RPCM}
In this subsection, Alternating Direction Method of Multipliers (ADMM) \cite{Dengw2017pmbadmm,lu2018unified} and a gradient descent (GD) algorithm \cite{Lij2015sdsn,Lij2017SAinDNN} are employed to solve the RPCM problem in Eq. \eqref{eq:RPCMmodel}. By replacing $\textbf{Z}-\text{diag}(\textbf{Z})$ by $\textbf{Z}$ to eliminate the constraint $\text{diag}(\textbf{Z})=0$ and introducing an auxiliary variable $\textbf{J}$ to deal with the multiple regularization, the Eq. \eqref{eq:RPCMmodel} is transformed into the following problem:
\begin{align}
\label{eq:RPCMequ}
\min_{\textbf{Z},\textbf{J},\textbf{E},\theta}\ & \|\textbf{Z}-f(\textbf{X};\theta)\|_{\text{F}}^2+ \overline{\alpha} R_1(\textbf{J})+\alpha R_2(\textbf{Z})+ \beta R_3(\textbf{E}), \nonumber \\
 \text{s.t.}\ & \textbf{X}=\textbf{X}\textbf{Z}+\textbf{E}, \ \textbf{Z}=\textbf{J}.
\end{align}
For convenience, we merge $\textbf{Z}$ and $\textbf{J}$ into a new variable $\mathbb{Z}$, and consider a matrix form of the problem in Eq. \eqref{eq:RPCMequ} as
\begin{align}
\label{eq:RPCMequmatrix}
\min_{\mathbb{Z},\mathbb{E},\theta} \ & \|\mathbb{Z}-f(\mathbb{X};\theta)\|_{\text{F}}^2+\widehat{\alpha}^{\text{T}} R_{\mathbb{Z}}(\mathbb{Z})+\widehat{\beta}^{\text{T}} R_{\mathbb{E}}(\mathbb{E}) \\
 \text{s.t.}  \ & \mathbb{X}=\mathbb{A}_1\mathbb{Z}+\mathbb{A}_2\mathbb{E}, \nonumber
\end{align}
where $\mathbb{X}$, $\mathbb{Z}$, $\mathbb{E}$, $\mathbb{A}_1$, $\mathbb{A}_2$,  $R_{\mathbb{Z}}(\mathbb{Z})$, $R_{\mathbb{E}}(\mathbb{E})$, $\widehat{\alpha}$ and $\widehat{\beta}$ are defined in Table \ref{tab:unifiedweights}. Note $\theta$ is essentially trained by $\|\textbf{Z}-f(\textbf{X};\theta)\|_{\text{F}}^2$, which is replaced by $\|\mathbb{Z}-f(\mathbb{X};\theta)\|_{\text{F}}^2$ for accordant notations. Then an augmented Lagrange function of the problem in Eq. \eqref{eq:RPCMequmatrix} is considered as:
\begin{align}
\mathcal{L}=& \|\mathbb{Z}-f(\mathbb{X};\theta)\|_{\text{F}}^2+\widehat{\alpha} R_{\mathbb{Z}}(\mathbb{Z})+\widehat{\beta} R_{\mathbb{E}}(\mathbb{E}) \nonumber \\
& +\frac{\mu}{2}\|\mathbb{X}-\mathbb{A}_1\mathbb{Z}-\mathbb{A}_2\mathbb{E}+\frac{\mathbb{Q}}{\mu}\|_{\text{F}}^2,
\label{eq:pldalm}
\end{align}
where $\mu$ is a penalty parameter and the Lagrange multiplier $\mathbb{Q}$ is provided in Table \ref{tab:unifiedweights}. To deal with the above problem, it iteratively updates $\theta$, $\{\mathbb{Z},\mathbb{E}\}$, $\mathbb{Q}$ and $\mu$ until convergence. By using a temporary variable $\overline{\mathbb{Z}}$, the key iterations are as follows.

%minimizing the function $\mathcal{L}$ with respect to $\theta$ as
\textbf{Updating $\theta^{k+1}$:} It is computed by training the parametric function with the initialized parameter $\theta^{k}$ to approximate to $\overline{\mathbb{Z}}^{k}$, that is, a subproblem\footnote{Here, the regularization (e.g., Frobenius norm or $\ell_2$ norm) of the weights is not provided for a simple expression in Eq. \eqref{eq:RPCMequ}. Generally, it can prevent the over-fitting of the weights.} is $\theta^{k+1}= \text{arg}\min_{\theta}\ \mathcal{L}_{\theta}^k$, where $\mathcal{L}_{\theta}^k= \|\overline{\mathbb{Z}}^{k}-f(\mathbb{X};\theta)\|_{\text{F}}^2$ is the loss. Gradient descent (GD) \cite{Lij2015sdsn,Lij2017SAinDNN} is a popular algorithm to solve this subproblem. With the initialization $\theta_{1}^{k}=\theta^k$, the update rule is
\begin{align}
\label{eq:solutiontheta}
\theta_{t+1}^{k}=\theta_{t}^{k}-\zeta \nabla\mathcal{L}_{\theta_{t}^{k}}^k,
\end{align}
where $\zeta$ is a learning rate, and $\nabla\mathcal{L}_{\theta_{t}^{k}}^k$ is the gradient of $\mathcal{L}_{\theta_{t}^{k}}^k$. After $T$ iterations, $\theta_{T}^{k}$ satisfies $\mathcal{L}_{\theta_{T}^{k}}^k<\epsilon$, where $\epsilon$ is an approximation error, and $\theta^{k+1}=\theta_{T}^{k}$.

\textbf{Calculating $\mathbb{Z}^k$:} It is computed by
\begin{align}
\label{eq:computezk}
\mathbb{Z}^{k}=f(\mathbb{X};\theta^{k+1}).
\end{align}

\textbf{Updating $\{\overline{\mathbb{Z}}^{k+1},\mathbb{E}^{k+1}\}$:} They are calculated by Jacobi-Proximal ADMM (JP-ADMM)  \cite{Dengw2017pmbadmm}. By adding the proximal terms $\|\mathbb{Z}-\mathbb{Z}^{k}\|_{\mathbb{P}_{\mathbb{Z}}}^2/2$ and $\|\mathbb{E}-\mathbb{E}^{k}\|_{\mathbb{P}_{\mathbb{E}}}^2/2$ into the Eq. \eqref{eq:pldalm}, the updated rules are described as
\begin{align}
\label{eq:solutionvzje}
\left\{\hspace{-0.1cm}\begin{array}{ccc}
\overline{\mathbb{Z}}^{k+1}=\text{arg}\min_{\mathbb{Z}}\ \widehat{\alpha} R_{\mathbb{Z}}(\mathbb{Z})+\frac{\mu^k}{2}\|\mathbb{X}-\mathbb{A}_1\mathbb{Z}-\mathbb{A}_2\mathbb{E}^k\\
-\frac{1}{\mu^k}\mathbb{Q}^k\|_{\text{F}}^2+\frac{1}{2}\|\mathbb{Z}-\mathbb{Z}^{k}\|_{\mathbb{P}_{\mathbb{Z}}}^2, \ \ \\
\mathbb{E}^{k+1}=\text{arg}\min_{\mathbb{E}}\ \widehat{\beta} R_{\mathbb{E}}(\mathbb{E})+  \frac{\mu^k}{2}\|\mathbb{X}-\mathbb{A}_1\mathbb{Z}^k-\mathbb{A}_2\mathbb{E}\\
-\frac{1}{\mu^k}\mathbb{Q}^k)\|_{\text{F}}^2 +\frac{1}{2}\|\mathbb{E}-\mathbb{E}^{k}\|_{\mathbb{P}_{\mathbb{E}}}^2,
\end{array}
\right.
\end{align}
where $\mathbb{P}_{\mathbb{Z}}\succeq0$ and $\mathbb{P}_{\mathbb{E}}\succeq0$ are symmetric and positive semi-definite matrices.

\textbf{Updating $\mathbb{Q}^{k+1}$:} the Lagrange multipliers are updated by:
\begin{align}
\label{eq:solutionq}
\mathbb{Q}^{k+1}=\mathbb{Q}^k+\gamma\mu^k(\mathbb{X}-\mathbb{A}_1\mathbb{Z}^k-\mathbb{A}_2\mathbb{E}^k),
\end{align}where $\gamma$ is a damping parameter.

\begin{algorithm}[tb]
\caption{RPCM via alternating JP-ADMM and GD.}
\begin{algorithmic}[1]
\State {\bfseries Input:} data matrix $\textbf{X}$ and parameters $\alpha, \beta, \gamma$.
\State {\bfseries Initialize:} $\theta$ is randomly initialized, $\mathbb{Z}^0=\overline{\mathbb{Z}}^0=0, \mathbb{E}^0=0, \mathbb{Q}^0=0, \zeta=0.01, \mu^0=10^{-2}, \mu_{max}=10^6, \rho=1.2, \epsilon_1=10^{-2}, \epsilon_2=10^{-4}$.
\State {\bfseries While} not converged \textbf{do}
\State \hspace{0.4cm} Repeat
\State \hspace{1.4cm} Update $\theta_{t+1}^{k}$ by Eq. \eqref{eq:solutiontheta},
\State \hspace{0.4cm} Until $\mathcal{L}_{\theta_{T}^{k}}^k<\epsilon_1$, $\theta^{k+1}=\theta_{T}^{k}$,
\State \hspace{0.4cm} Update $\mathbb{Z}^{k}$ by Eq. \eqref{eq:computezk},
\State \hspace{0.4cm} Update $\{\overline{\mathbb{Z}}^{k+1},\mathbb{E}^{k+1}\}$ by Eq. \eqref{eq:solutionvzje},
\State \hspace{0.4cm} Update $\mathbb{Q}^{k+1}$ by Eq. \eqref{eq:solutionq},
\State \hspace{0.4cm} Update the parameter $\mu^k=\min\{\rho\mu^k,\mu_{max}\}$,
\State \hspace{0.4cm} Check: $\|\mathbb{X}-\mathbb{A}_1\mathbb{Z}^k-\mathbb{A}_2\mathbb{E}^k\|_{\text{F}}^2<\epsilon_2$.
\State {\bfseries End}
\State {\bfseries Return} $\theta$, $\mathbb{Z}$ and $\mathbb{E}$.
\end{algorithmic}
\label{alg:RPCM}
\vskip -0.0in
\end{algorithm}

Overall, the procedure iterates above three steps and $\mu$ until a stopping is satisfied. \textbf{Algorithm \ref{alg:RPCM}} summarizes the whole optimization procedure. Note that in practice the lines 4-6 can be moved to line 12. This leads to less optimization time since it only needs one time to train the neural network. Give new data points, their ideal codes are easily calculated by the parametric function in Eq. \eqref{eq:computezk} with \emph{linear} time to the number of data points.

\subsection{Convergence Analysis and Time Complexity} % Due to the nonlinear parametric function and the multi-variables minimization, it is a challenge to theoretically prove the convergence. Inspired by the proof method \cite{Lij2018pcm}
% where $\textbf{I}_d$ and $\textbf{I}_d$ are identity matrices of size $d\times d$ and $D\times D$, respectively.
\textbf{Convergence Analysis.} The convergence analysis of RPCM in \textbf{Algorithm \ref{alg:RPCM}} is discussed in this subsection. RPCM without steps 4-6 is a standard JP-ADMM \cite{Dengw2017pmbadmm}, and by choosing $0<\gamma<2$, $\mathbb{P}_{\mathbb{Z}}=\tau_{\mathbb{Z}}\textbf{I}-\mu^k\mathbb{A}_1^{\text{T}}\mathbb{A}_1$ and $\mathbb{P}_{\mathbb{E}}=\tau_{\mathbb{E}}\textbf{I}-\mu^k\mathbb{A}_2^{\text{T}}\mathbb{A}_2$, its practical condition is
\begin{align}
\label{eq:Theorem1condition2}
\left\{\hspace{-0.1cm}\begin{array}{ccc}
\tau_{\mathbb{Z}}>\frac{2}{2-\gamma}\mu^k\|\mathbb{A}_1\|^2,\\
\tau_{\mathbb{E}}>\frac{2}{2-\gamma}\mu^k\|\mathbb{A}_2\|^2. \\
\end{array}
\right.
\end{align}Following the approximation theorem\footnote{$f(\mathbb{X};\theta)$ with enough hidden units can approximate uniformly $\overline{\mathbb{Z}}^{k}$ on compact \cite{Ripley1996}.} of the neural network and the convergence of GD or SGD \cite{Lij2015sdsn,Lij2017SAinDNN,Reddi2016svrno}, the steps 4-6 are used to learn the neural parametric function such that $\mathcal{L}_{\theta}^k= \|\overline{\mathbb{Z}}^{k}-f(\mathbb{X};\theta)\|_{\text{F}}^2<\epsilon_1$ with enough small $\epsilon_1$. Under $\mathcal{L}_{\theta}^k<\epsilon_1$ , we empirically verify that RPCM is still convergent in Fig. \ref{fig:convergence}. Notation that the steps 4-6 can be moved to end in the Algorithm \ref{alg:RPCM} for reducing the training time. %This means that independent J-ADMM and GD also can make RPCM convergent.

\begin{table}[t]
\linespread{0.75}
\vskip -0.0in
\setlength{\tabcolsep}{1.3pt}
\renewcommand{\arraystretch}{1.3}
\caption{Complexities of the classical coding methods and our RPCM.}
\vskip -0.15in
\label{tab:computcompex}
\begin{center}
%\scriptsize \footnotesize,
\begin{tabular}{c|c|c}
\Xhline{1.2pt}
Mehtods                 & training time                    & coding time                     \\
\hline
RPCM$_{\ell_1\hspace{-0.05cm}+\hspace{-0.01cm}\text{F}^2}$ &$\mathcal{O}(t_2(t_1\hspace{-0.1cm}+\hspace{-0.1cm}d\hspace{-0.1cm}+\hspace{-0.1cm}1)n^3)$ & \multirow{4}{*}{ $\mathcal{O}\left((\sum_{i=2}^l\iota_i\iota_{i-1})m\right)$}    \\
RPCM$_{\ell_1}$               &$\mathcal{O}(t_2((t_1\hspace{-0.1cm}+\hspace{-0.1cm}D)n^3))$ &    \\
RPCM$_{\ast}$                 &$\mathcal{O}(t_2((t_1\hspace{-0.1cm}+\hspace{-0.1cm}1)n^3\hspace{-0.1cm}+\hspace{-0.1cm}dn^2))$ &    \\
RPCM$_{\text{F}^2}$            &  $\mathcal{O}((t_1\hspace{-0.1cm}+\hspace{-0.1cm}1)n^3)$ &   \\
%RPCM$_{\ell_1\hspace{-0.05cm}+\hspace{-0.01cm}\ast}$       & $\mathcal{O}(T_2((T_1\hspace{-0.1cm}+\hspace{-0.1cm}D\hspace{-0.1cm}+\hspace{-0.1cm}1)N^3\hspace{-0.1cm}+\hspace{-0.1cm}DN^2))$ &    \\
\hline
ENSC \cite{Panagakis2014ENSC}& \multirow{4}{*}{$\mathcal{O}(0)$} & $\mathcal{O}(t_2((d\hspace{-0.1cm}+\hspace{-0.1cm}1)m^3))$          \\
SSC \cite{Elhamifar2013}   &                  & $\mathcal{O}(t_2dm^3)$   \\
LRR \cite{Liugc2013}       &                  & $\mathcal{O}(t_2(dm^2\hspace{-0.1cm}+\hspace{-0.1cm}m^3))$      \\
LSR \cite{Lucy2012}        &                    & $\mathcal{O}(m^3)$                \\
%LRSSC \cite{Wang2013lrssc} &                  & $\mathcal{O}(T_2(DM^2\hspace{-0.1cm}+\hspace{-0.1cm}(D\hspace{-0.1cm}+\hspace{-0.1cm}1)M^3))$     \\
\Xhline{1.2pt}
\end{tabular}%}
\\
\end{center}
\begin{flushleft}
Notation: $d$:$\sharp$ of dimension; $n$:$\sharp$ of representatives; $m$:$\sharp$ of all data points;
$l$: $\sharp$ of hidden layers; $\iota_i$: $\sharp$ of hidden units of $i$th-layer;
$t_1$: training epochs; $t_2$: $\sharp$ of iterations.
\end{flushleft}
\vskip -0.2in
\end{table}
% \multicolumn{3}{|l|}{$n$:$\sharp$ of samples; $d$:$\sharp$ of dimensionality of sample; $m$:$\sharp$ of the selected samples; $h$:$\sharp$} \\
% \multicolumn{3}{|l|}{ of hidden units of non-linear function; $t_1, t_2, T_m$:$\sharp$ of the number of iterations in}\\
% \multicolumn{3}{|l|}{ $l_1$ solver, rank-minimizer, training the non-linear function; $\mathcal{O}(0)$: no training time.}\\

\textbf{Time Complexity.}
The key computation is steps 4-6 and 8 in Algorithm \ref{alg:RPCM}. The complexity of steps 4-6 is $\mathcal{O}(t_1n^3)$, where $t_1$ is the number of training epochs. Since the weights of the top layer in the neural network can be formulated as a convex problem and it has a closed-form solution \cite{Lij2015sdsn}, $t_1$ is less than $5$ epochs in this paper. In the step 8, there are different complexities for the different choices of $R(\textbf{Z})$ shown in Table \ref{tab:fivenorms}. The basic complexities of  matrix inversion for the square of Frobenius norm, LASSO for $\ell_1$ norm and singular value decomposition (SVD) for unclear norm are $\mathcal{O}(n^3)$, $\mathcal{O}(dn^3)$ and $\mathcal{O}(dn^2\hspace{-0.05cm}+\hspace{-0.05cm}n^3)$, respectively. Considering the number of iterations $t_2$ needed to converge, thus, the overall computational complexities of RPCM$_{\text{F}^2}$, RPCM$_{\ell_1}$, RPCM$_{\ast}$, and RPCM$_{\ell_1\hspace{-0.05cm}+\hspace{-0.01cm}\text{F}^2}$ are $\mathcal{O}((t_1+1)n^3)$, $\mathcal{O}(t_2((t_1+d)n^3))$, $\mathcal{O}(t_2((t_1+1)n^3+dn^2))$, and $\mathcal{O}(t_2(t_1+d+1)n^3)$, respectively. The compared complexities of RPCM and the classical coding methods (e.g. SSC, LRR, LSR, and ENSC) are shown in Table \ref{tab:computcompex}. Since the number of representatives is much less than the number of all data points, i.e., $n\ll m$, the training complexity can be ignored. Moreover, the coding complexity $\mathcal{O}\left((\sum_{i=2}^l\iota_i\iota_{i-1})m\right)$ is \emph{linear} in terms of $m$. Compared to the classical coding methods, thus, our RPCM is easy to compute the codes in large-scale datasets.

%Note that although RPCM spends time to train the parametric function (e.g., neural network), the steps 4-6 can be moved to end in Algorithm \ref{alg:RPCM} for saving the training iteration time.

\emph{\textbf{Remark 4.}} According to the complexities of the classical coding methods, they are infeasible when $m$ is larger than one million. Although some recent works (e.g., SSC-OPM \cite{Youc2016sssc}, EnSC\footnote{EnSC \cite{Youc2016ensc} is different from ENSC \cite{Panagakis2014ENSC} as EnSC is a scalable method to optimize the elastic-net codes for subspace clustering.} \cite{Youc2016ensc}) are to accelerate the coding methods, they are still too expensive due to the million dimensions based on the self-expressive dictionary. To implement the coding methods, the representatives are also chosen as their self-dictionary, that is, the Eq. \eqref{eq:RPCMmodel} without the learning term, in the latter experiments. Moreover, the computational time are still much expensive than RPCM in Table \ref{tab:timeonmnist}.

\section{Contractive Subspace Recovery Theory}
\label{sec:theoreticalanalysis}
To cluster millions of data points, the success of the LeaSC paradigm is based on an underlying assumption that the parametric function learned from the representatives can contrastively calculate a subspace representation of every data point. In this section, we investigate that the parametric function can calculate more contractive representations than the traditional subspace clustering methods.

%it is a natural practicable strategy to choose \emph{representatives} as the self-dictionary in the classical coding models of the popular subspace clustering methods. Compared to these methods,
%Note $K$-means \cite{Caid2015} or DS3 \cite{Elhamifar2016subset} is often employed to choose $N$ \emph{representatives} $\textbf{X}=[\textbf{X}_1,\cdots,\textbf{X}_k]\in{\rm I\!R}^{D\times N} (N<M)$ to well describe $\textbf{Y}$. Specifically, Each subspace $\mathcal{X}_i$ contains $M_i$ data points , which are well described by $N_i$ \emph{representatives} with $\sum_{i}N_i=N$.
Formally, recalling the i.i.d. assumption, the huge data matrix $\textbf{Y}\in{\rm I\!R}^{d\times m}$ (i.e., $m>1,000,000$) and the representative matrix $\textbf{X}\in{\rm I\!R}^{d\times n} (n<m)$ are i.i.d. sampled from the unknown distribution $p(X)$ in the union of subspaces $\{\mathcal{X}_i\}_{i=1}^s$.  Let $\textbf{Y}_i\in{\rm I\!R}^{d\times m_i}$ and $\textbf{X}_i\in{\rm I\!R}^{d\times n_i}$ denote the submatrix in $\textbf{Y}$ and the representative submatrix in $\textbf{X}$ that belongs to $\mathcal{X}_i$ with $\sum_{i}m_i=m$ and $\sum_{i}n_i=n$. Without loss of generality, let $\textbf{Y}=[\textbf{Y}_1,\cdots,\textbf{Y}_s]$ and $\textbf{X}=[\textbf{X}_1,\cdots,\textbf{X}_s]$ be ordered. To show the representative ability of $\textbf{X}$, we first introduce \emph{representative-radius} of $(\textbf{X}_i,\textbf{Y}_i)$ as follows:

\textbf{Definition 1.} \emph{We let} $\textbf{Y}_i=[\textbf{Y}_{i1},\cdots,\textbf{Y}_{in_i}]$ \emph{with} $\textbf{Y}_{ij}\in{\rm I\!R}^{d\times m_{ij}}$ \emph{and} $\textbf{X}_i=[\textbf{x}_{i1},\cdots,\textbf{x}_{in_i}]$\emph{, and assume that a representative} $\textbf{x}_{ij}$ \emph{well represents the submatrix} $\textbf{Y}_{ij}$ with $\sum_{j=1}^{n_i}m_{ij}=m_{i}$\emph{. We define the representative-radius associated with} $(\textbf{x}_{ij},\textbf{Y}_{ij})$ \emph{as:}
\begin{align}
\label{eq:representative-radius}
\rho_{ij}=\max_{\textbf{y}\in\textbf{Y}_{ij}}d(\textbf{x}_{ij},\textbf{y}),
\end{align}
\emph{where} $d(\textbf{x}_{ij},\textbf{y})$ \emph{is a dissimilarity between} $\textbf{x}_{ij}$ \emph{and} $\textbf{y}$ \emph{(e.g., encoding error or Euclidean distance). The representative-radiuses associated with} $(\textbf{X}_i,\textbf{Y}_i)$ \emph{and} $(\textbf{X},\textbf{Y})$ \emph{are defined as} $\rho_{i}=\max_j\rho_{ij}$ \emph{and} $\rho=\max_i\rho_{i}$\emph{, respectively.}

Based on the radius $\rho$ in Definition 1, we denote $\mathcal{N}_{\textbf{X}}^\rho$ as the neighborhood of the representatives $\textbf{X}$ with the radius $\rho$. With the radius $\rho$ and the suitable number of the representatives $\textbf{X}$, all data points $\textbf{Y}$ will fall in $\mathcal{N}_{\textbf{X}}^\rho$.

\subsection{Contraction Analysis}
In this subsection, we analyze two contractions of the parametric function $f(\bullet;\theta)$, where $\theta$ is the solution of our PCM model. First, the parametric function can calculate contractive representations. Specifically, the representations $f(\textbf{Y};\theta)$ can approximate to the subspace representations $\textbf{Z}=f(\textbf{X};\theta)$ with a upper bound which is dependent on $\rho$.

\begin{figure}[t]
\vskip -0.0in
\begin{center}
\centerline{\includegraphics[width=0.65\columnwidth]{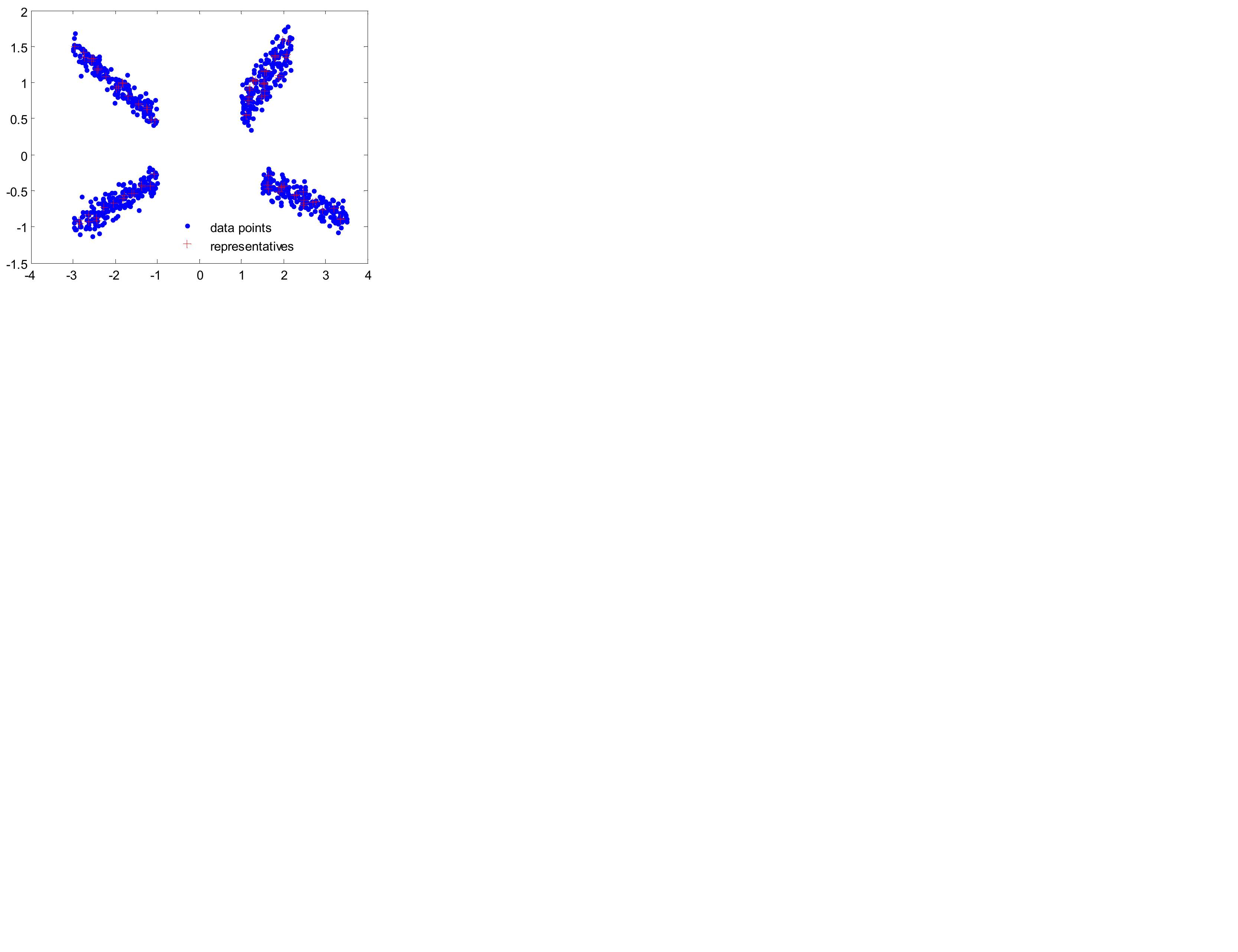}}
\vskip -0.1in
\caption{An artificial dataset including 800 data points (blue dots) with two dimensions sampled from 4 linear subspaces with small gaussian noises. The representatives (red pluses) found by DS3 \cite{Elhamifar2016subset} for our LeaSC paradigm.}
\label{fig:exampledata}
\end{center}
\vskip -0.1in
\end{figure}

\begin{figure*}[ht]
\vskip -0.0in
\begin{center}
\centerline{\includegraphics[width=1.8\columnwidth]{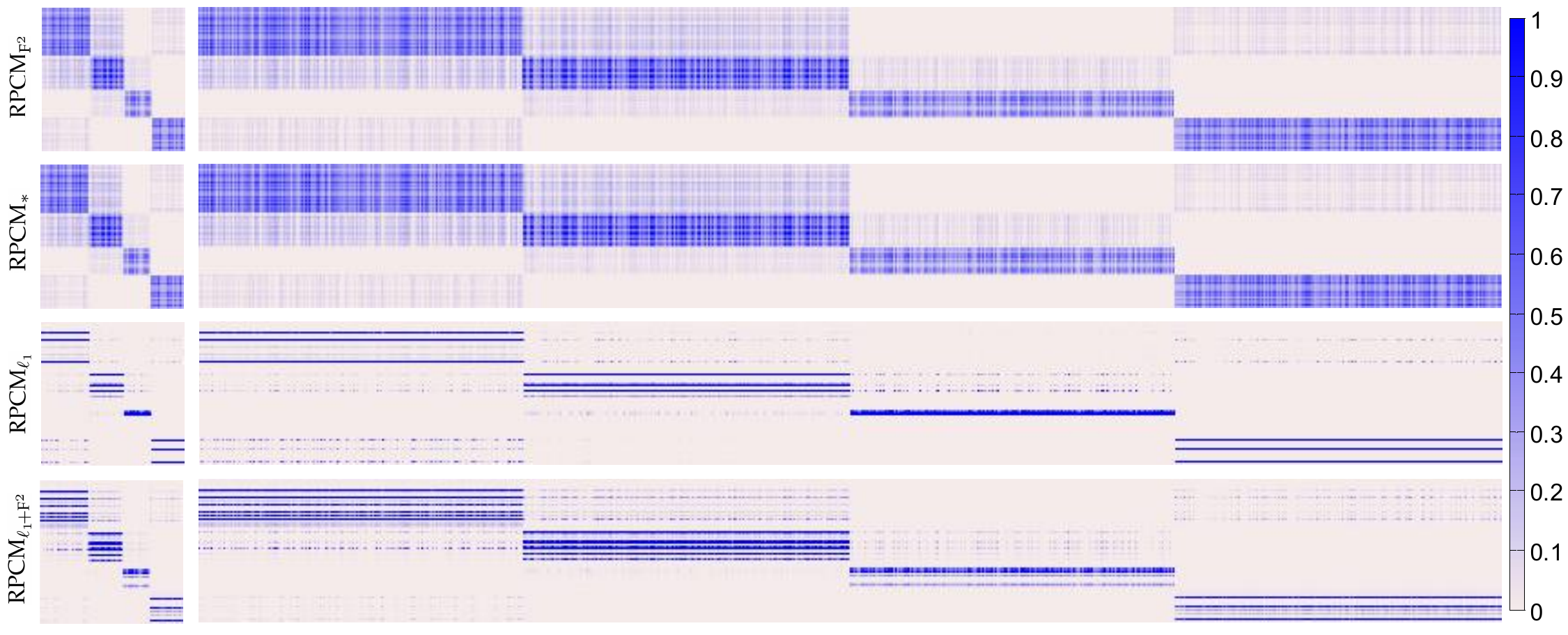}}
\vskip -0.1in
\caption{Visualizations of the low-dimensional codes obtained by RPCM on an artificial dataset: from up to down, RPCM$_{\text{F}^2}$, RPCM$_{\ast}$, RPCM$_{\ell_1}$ and RPCM$_{\ell_1\hspace{-0.05cm}+\hspace{-0.01cm}\text{F}^2}$; from left to right, the codes of the representatives, and the codes of all data points.}
\label{fig:example-2}
\end{center}
\vskip -0.1in
\end{figure*}

\begin{figure*}[ht]
\vskip -0.0in
\begin{center}
\centerline{\includegraphics[width=1.8\columnwidth]{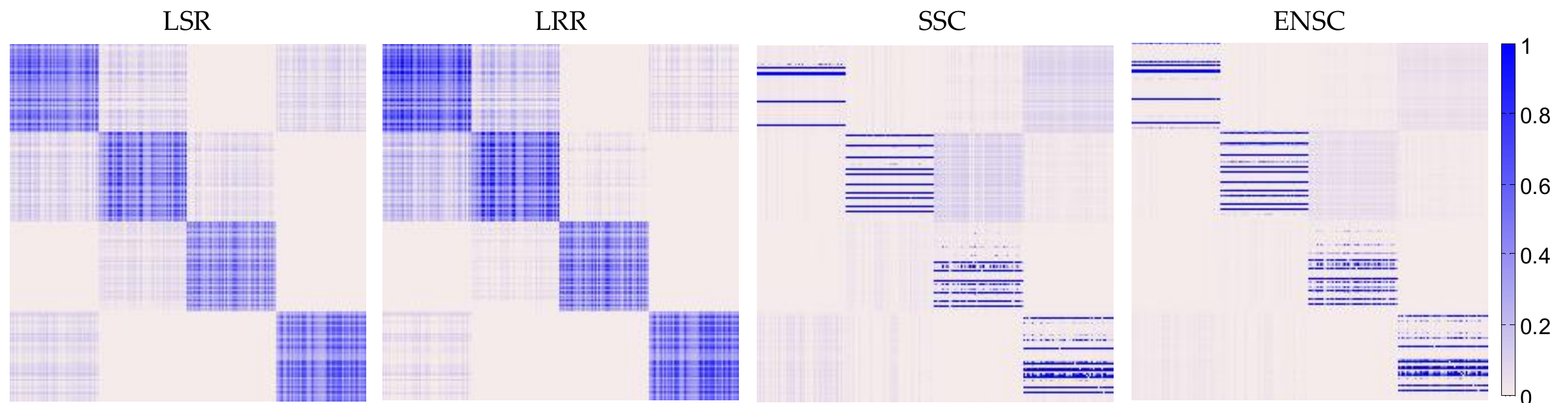}}
\vskip -0.1in
\caption{Visualizations of the high-dimensional codes obtained by the popular subspace clustering methods on an artificial dataset: from up to down, LSR \cite{Lucy2012}, LRR \cite{Liugc2013}, SSC \cite{Elhamifar2013} and ENSC \cite{Panagakis2014ENSC}.}
\label{fig:example-1}
\end{center}
\vskip -0.2in
\end{figure*}

%by using the parametric function $f(\bullet;\theta)$. Specifically, $f(\textbf{Y};\theta)$ is close to $\textbf{Z}=f(\textbf{X};\theta)$, where $\textbf{Z}$ and $\theta$ are the solutions of our PCM model. % with $\textbf{X}$. %in the Eq. \eqref{eq:hugedatacodingmodel}.

%To solve the LSSC problem, FRC trains the function \eqref{eq:project} to approximate the codes learned by \eqref{eq:rcmmatrix}. In this subsection, we show an approximation condition and a grouping effect to verify that FRC is fit for large-scale subspace clustering.
%\textbf{Approximate the subspace representations by FRC.} A sufficient condition is to show that FRC can effectively approximate the subspace representations.
%$\|\bullet\|_{c}$ \emph{denote the Frobenius or unclear norm, and let}
\textbf{Theorem 1.} (The proof is provided in the appendix B.) \emph{Assume the huge data matrix} $\textbf{Y}$ \emph{and the representative matrix} $\textbf{X}$ \emph{are i.i.d. sampled from the union of subspaces} $\{\mathcal{X}_i\}_{i=1}^s$\emph{. Let} $\mathcal{N}_{\textbf{X}}^\rho$ \emph{be the neighborhood of} $\textbf{X}$ \emph{with the radius} $\rho$. \emph{For any} $\textbf{y}\in\textbf{Y}_{ij}\subset\textbf{Y}_i\subset\textbf{Y}\subset\mathcal{N}_{\textbf{X}}^\rho$ \emph{and} $\textbf{x}_{ij}\in\textbf{X}_i\subset\textbf{X}$ $(1\leq j\leq N_i, 1\leq i\leq s)$, \emph{a upper bound of the distance between} $f(\textbf{x}_{ij};\theta)$ \emph{and} $f(\textbf{y};\theta)$ \emph{are}
 \begin{align}
\label{eq:contractionresult}
&\|f(\textbf{x}_{ij};\theta)-f(\textbf{y};\theta)\| \nonumber \\
\leq &
\left\{
   \begin{array}{ll}
   O(\rho), & \hbox{if $\|\frac{\partial f(\textbf{x}_{ij};\theta)}{\partial \textbf{x}_{ij}}\|_{\textbf{F}}\leq \rho$,}\\
   \|\frac{\partial f(\textbf{x}_{ij};\theta)}{\partial \textbf{x}_{ij}}\|_{\textbf{F}}\rho+O(\rho), & \hbox{ otherwise.}
   \end{array}
   \right. ,
\end{align}\emph{where} $\theta$ \emph{is a solution of the PCM model, and} $O(\rho)$ \emph{is an infinitesimal of higher order than} $\rho$.

%\emph{where} $\textbf{z}_{ij}^\star\in\textbf{Z}_i^\star\subset\textbf{Z}^\star$ \emph{and} $\textbf{z}^{\textbf{y}}\in\textbf{Z}^{\textbf{Y}_{ij}}\subset\textbf{Z}^{\textbf{Y}_{i}}\subset\textbf{Z}^{\textbf{Y}}$ \emph{are the representations respectively solved in Eq. \eqref{eq:representivecodingmodel} and Eq. \eqref{eq:hugedatacodingmodel}, and} $\textbf{X}^\dag$ \emph{is the pseudoinverse of} $\textbf{X}$.
%, that is, the representations $f(\textbf{y};\theta)$ can contractively approximate to the subspace representations $f(\textbf{x}_{ij};\theta)$.

Theorem 1 implies the bounded contraction of the learned parametric function $f(\bullet;\theta)$. When the representative-radius $\rho$ is small, the upper bound is $\|\frac{\partial f(\textbf{x}_{ij};\theta)}{\partial \textbf{x}_{ij}}\|_{\textbf{F}}\rho$. In fact, when the representatives are enough, it is easy to get a small $\rho$ which leads to a low upper bound. Moreover, if $\|\frac{\partial f(\textbf{x}_{ij};\theta)}{\partial \textbf{x}_{ij}}\|_{\textbf{F}}\leq \rho$ is satisfied, then the upper bound is an infinitesimal of higher order than $\rho$. This shows that $f(\bullet;\theta)$ is a beneficial computational function for the subspace clustering tasks since it has a less invariance of the subspace representation for all data points $\textbf{Y}$ fallen in $\mathcal{N}_{\textbf{X}}^\rho$.

Second, the parametric function can obtain the contractive dimension. Based on the SE property \cite{Elhamifar2013}, the number of the dimension of the representation is the same to the number of all data points in the traditional subspace clustering methods (e.g., LSR \cite{Lucy2012}, LRR \cite{Liugc2013}, SSC \cite{Elhamifar2013} and ENSC \cite{Panagakis2014ENSC}). When we face the huge datasets, they have very high dimension. Compared to these methods, the number of the output of the parametric function is the same to the number of the representatives, which is less than the number of all data points. Thus, the parametric function can well calculate the low-dimensional representations. In the next subsection \ref{sec:ansmallexperiment}, an experiment verifies the bounded contractive representations $f(\bullet;\theta)$ and its low-dimension.

%To any other data point in $\mathcal{N}_{\textbf{X}}^\rho$, in addition, $f(\bullet;\theta)$ still fast compute its contractive representation so that our RPCM model can deal with millions of data points.
%\emph{\textbf{Remark 5:}} Based on the dictionary $\textbf{X}$, the representation $\textbf{z}^{\textbf{y}}$ can well reconstruct the data point $\textbf{y}$, while $f(\textbf{y};\theta)$ approximately regenerates the representative of $\textbf{y}$.

\emph{\textbf{Remark 5:}} In the popular subspace clustering theories, it is well-known that the group effect \cite{Lucy2012}, the sparse recovery \cite{Elhamifar2013} and the low-rank recovery \cite{Liugc2013} have been proposed for LSR, SSC, and LRR. The theoretical guarantee of ENSC \cite{Panagakis2014ENSC} has also been studied by mixing the group effect with the sparse recovery. By combining these theories with our contractive guarantee, it is easy to obtain contractive group effect, contractive sparse/low-rank recovery and their mixed variants. We omit them in this paper.

% The condition \eqref{eq:condition} is easily satisfied because $f(\cdot;\theta)$ is learned from the FRC problem \eqref{eq:frc}. Hence, the projective codes can approximate to the subspace representations by a first-order accuracy as long as $\textbf{U}$ drawn from $\textbf{X}$ is belonging to the neighbourhood of $\textbf{Y}$.

\subsection{An Experiment on a Synthetic Dataset}
\label{sec:ansmallexperiment}
This subsection, we give an intuitive experiment to verify the effectiveness of RPCM in the LeaSC paradigm. We generate 800 data points with two dimensions, $\textbf{Y}=[\textbf{Y}_1,\cdots,\textbf{Y}_4]$ from 4 linear subspaces with small gaussian noises. We sample 200 data points for each subspace, and they are plotted in Fig. \ref{fig:exampledata}. Following our LeaSC paradigm, we found 88 representatives (red pluses), $\textbf{X}=[\textbf{X}_1,\cdots,\textbf{X}_4]$, shown in Fig. \ref{fig:exampledata} by DS3 \cite{Elhamifar2016subset}. The representatives are used to train a parametric function, a three-layer neural network with 50 hidden units, by our RPCM with different regularizations in Eq. \eqref{eq:RPCMmodel}. The representations of $\textbf{X}$ are plotted in the first column of Fig. \ref{fig:example-2}, and the second column of Fig. \ref{fig:example-2} shows the low-dimensional representations of all data points, which are fast calculated by the learned neural network. We observed that the learned neural network well contractively maps all data points into the low-dimensional structures. Compared to the traditional subspace clustering methods (e.g., LSR \cite{Lucy2012}, LRR \cite{Liugc2013}, SSC \cite{Elhamifar2013} and ENSC \cite{Panagakis2014ENSC}), RPCM only has the 88 dimensions, which is much lower than 800 dimensions shown in Fig. \ref{fig:example-1}. Note that since we generate two dimensional data points for visualization, the representations learned by RPCM$_{\text{F}^2}$ is similar to RPCM$_{\ast}$ and the representations obtained by LSR is also similar to LRR.

\section{Experiments}
\label{sec:experiments}
In this section, we conduct several experiments to verify the efficiency and effectiveness of our LeaSC. Firstly, six real-world datasets including two small datasets and four large datasets are described. Secondly, we introduce mulitple state-of-the-art methods and evaluation metrics. Finally, we show the clustering results.

\begin{figure}[t]
\vskip -0.0in
\begin{center}
\centerline{\includegraphics[width=0.85\columnwidth]{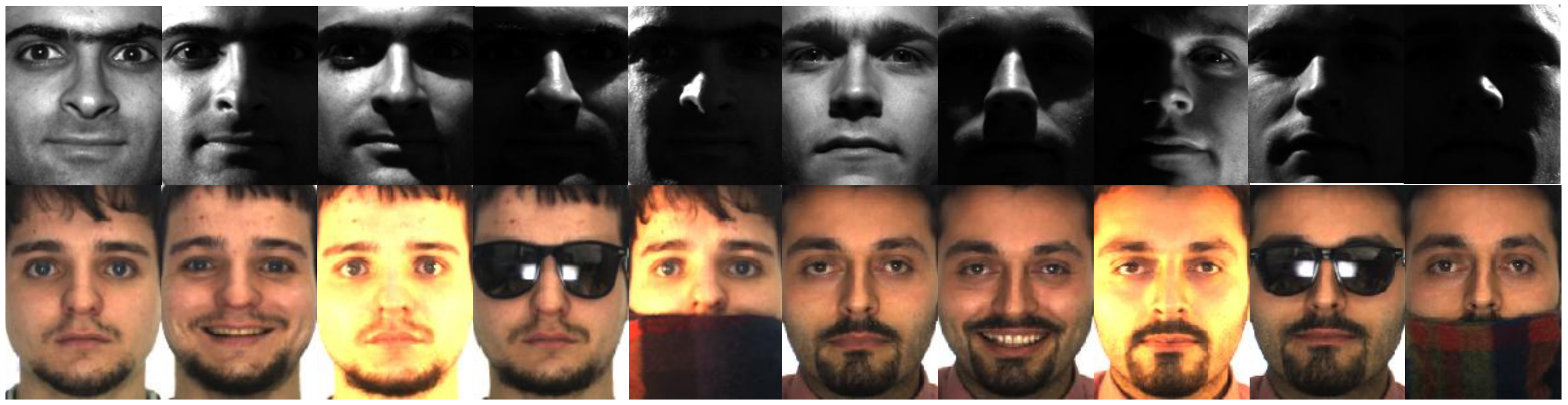}}
\vskip -0.1in
\caption{Examples of Extended-YaleB (up) and AR (down).}
\label{fig:imageExtended-YaleB-AR}
\end{center}
\vskip -0.2in
\end{figure}

\begin{figure}[t]
\vskip -0.1in
\begin{center}
\centerline{\includegraphics[width=0.85\columnwidth]{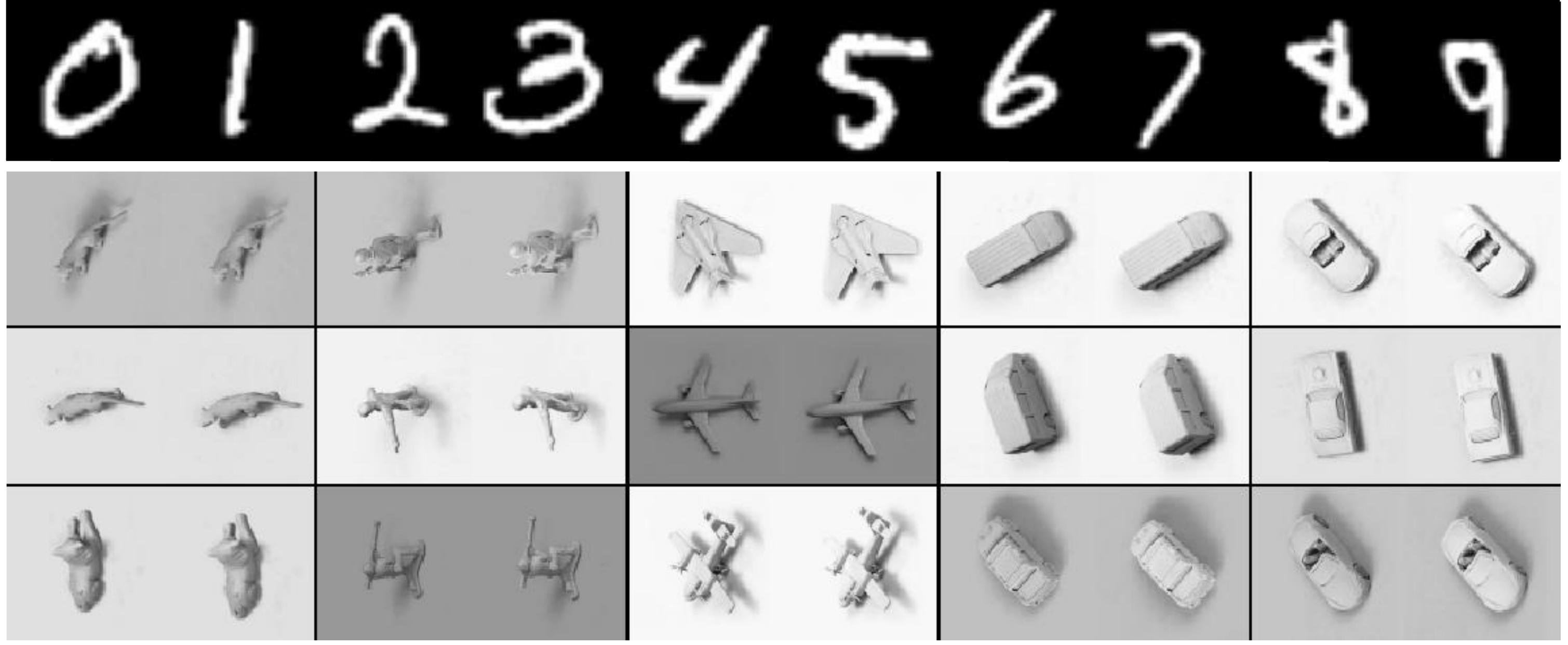}}
\vskip -0.1in
\caption{MNIST (up) contains some greyscale images of handwritten digits 0-9, and NORB (down) consists of 3D objects: from left to right, four-legged animals, human figures, airplanes, trucks, and cars.}
\label{fig:imagemnistnorb}
\end{center}
\vskip -0.2in
\end{figure}

\begin{figure}[t]
\vskip -0.0in
\begin{center}
\centerline{\includegraphics[width=0.85\columnwidth]{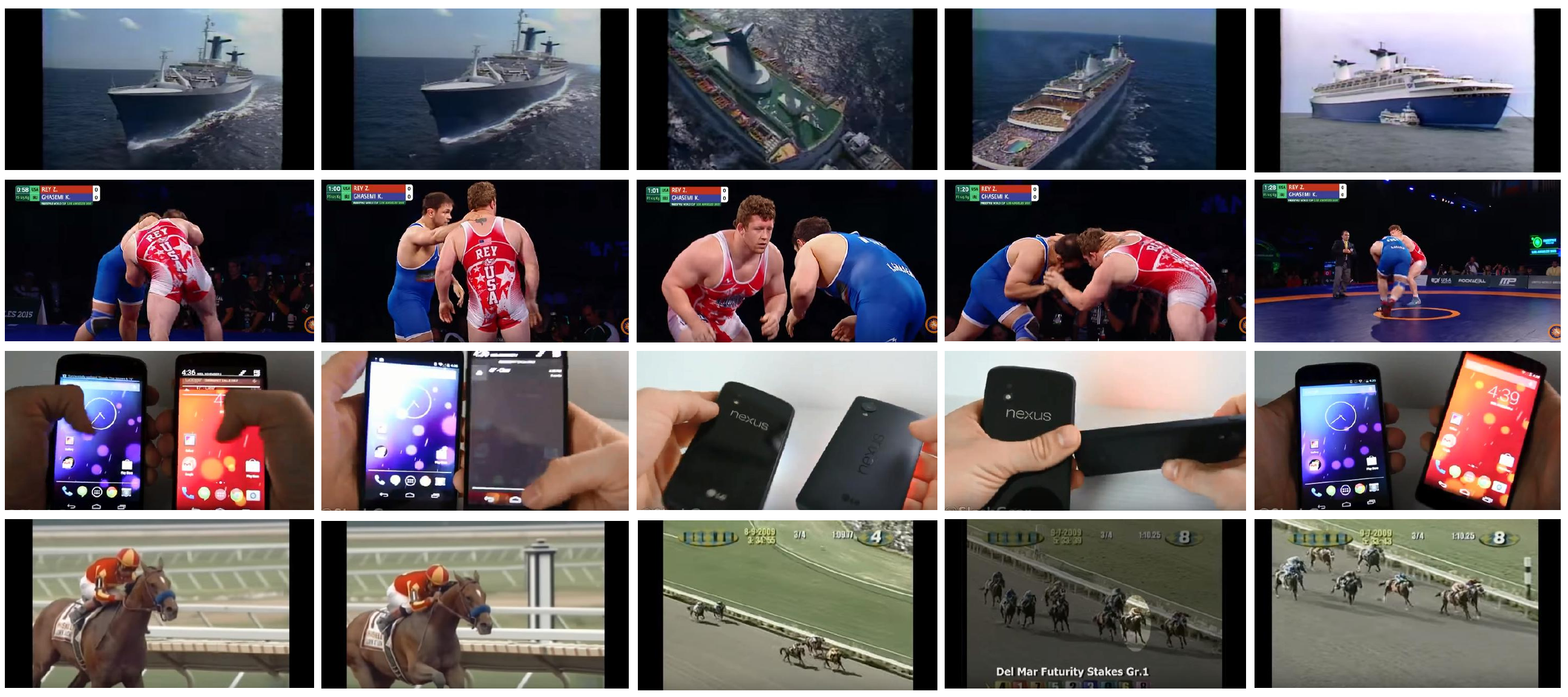}}
\vskip -0.1in
\caption{Four videos in YouTube-8M-1M: from up to down, Boat, Wrestling, Smartphone, and Racing.}
\label{fig:videosYouTube-8M-1M}
\end{center}
\vskip -0.2in
\end{figure}

\begin{table}[t]
%\linespread{0.75}
%\renewcommand{\arraystretch}{1.3}
\vskip -0.0in
\caption{Datasets.}
\vskip -0.15in
\label{tab:databases}
%\small{
\begin{center}
\scalebox{1}{
\begin{tabular}{c|c|c|c}
\Xhline{1.2pt}
 Datasets      & $\#$ samples & $\#$ dimensions & $\#$ classes  \\
\hline
Extended-YaleB     & 2,414       & 167    & 38 \\
AR             & 2,600        & 114       & 100    \\
NORB           & 48,600       & 2,048     & 6      \\
MNIST          & 70,000       & 3,472       & 10     \\
MNIST8M-1M  & 1,000,000    & 3,472     & 10     \\
YouTube8M-1M  & 1,000,000    & 1,024     & 100    \\
\Xhline{1.2pt}
\end{tabular}}
\end{center}
\vskip -0.2in
\end{table}
\subsection{Datasets}
We describe two kinds of datasets: two small datasets and three large datasets as shown in Table \ref{tab:databases}.

\textbf{Small datasets.} \emph{Extended-YaleB} includes 2,414 frontal face images with 38 people. There are about 64 images for each person.  Extended-YaleB\footnote{http://vision.ucsd.edu/~leekc/ExtYaleDatabase/ExtYaleB.html}, \emph{AR}\footnote{http://www2.ece.ohio-state.edu/~aleix/ARdatabase.html} contains over 4,000 images with 126 people. We select 50 male and 50 female persons with 26 face images to create a subset consisting of 2,600 images. The original images of these two datasets were cropped and normalized to $48\times42$ and $55\times40$ pixels, and were reduced to 167 and 114 dimensions by PCA, respectively. Some examples are shown in Fig. \ref{fig:imageExtended-YaleB-AR}.

\textbf{Large datasets.} \emph{NORB}\footnote{http://www.cs.nyu.edu/~ylclab/data/norb-v1.0-small/} has 48,600 $2\times 96 \times 96$ images of 3D object toys belonging to 5 generic categories: airplanes, cars, trucks, four-legged animals and human figures. The original images were subsampled to $2 \times 32 \times 32$. \emph{MNIST}\footnote{http://yann.lecun.com/exdb/mnist/} consists of 70,000 samples with $28\times28$ pixel images of handwritten digits 0-9. Following the settings \cite{Youc2016ensc}, we compute the feature vectors with 3,472 dimensions by using a scattering convolution network \cite{Brunaj2013}. \emph{MNIST8M-1M} randomly selects 1 million samples from MNIST8M\footnote{http://leon.bottou.org/projects/infimnist} constructed by extending MNIST to produce 8 million images. Similar to MNIST, we also compute the 3,472 dimensional feature vectors. \emph{YouTube8M-1M} randomly selects 1 million video-level features with 100 classes from YouTube8M\footnote{https://research.google.com/youtube8m/download.html}, where the size of each feature is 1,024. Some videos are shown in Fig. \ref{fig:videosYouTube-8M-1M}.
%Some examples of NORB and MNIST are shown in Fig. \ref{fig:imagemnistnorb}.

\begin{figure*}[ht]
\vskip -0.0in
\begin{center}
\centerline{\includegraphics[width=1.85\columnwidth]{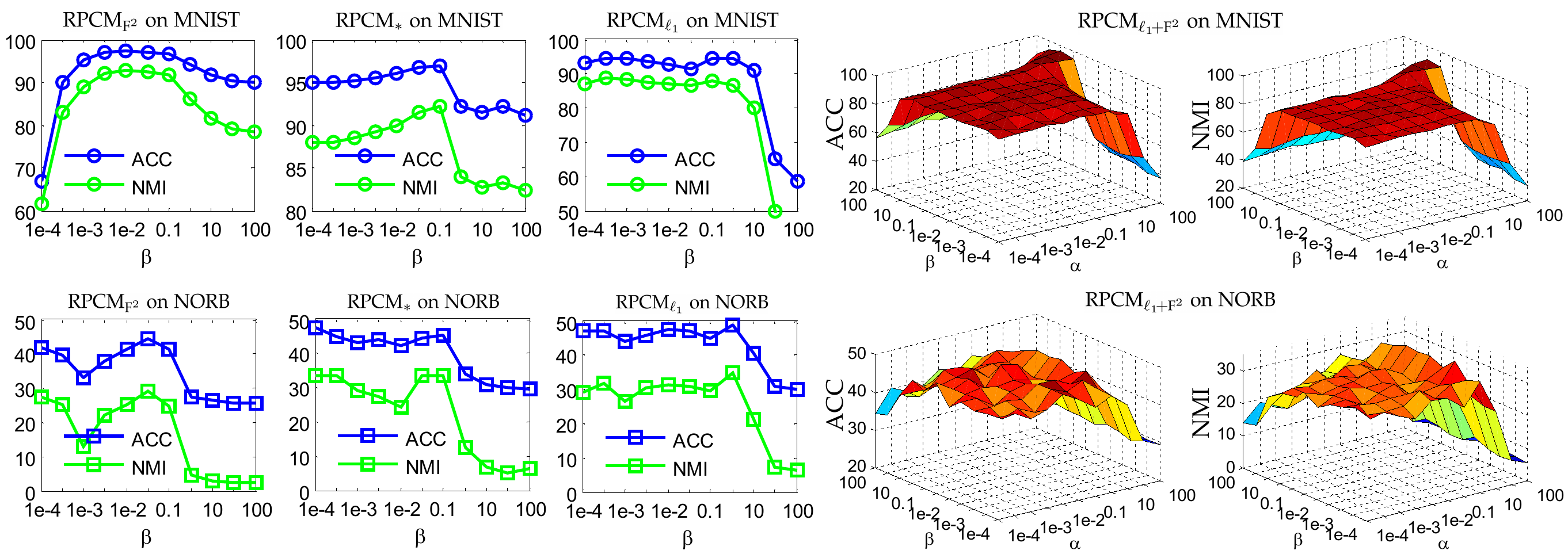}}
\vskip -0.1in
\caption{ACC $(\%)$ and NMI $(\%)$ with different parameters $\alpha$ and $\beta$, which are selected from $\{1e^{-4},5e^{-4},1e^{-3},5e^{-3},1e^{-2},$ $5e^{-2}, 0.1,1,10, 50, 100\}$. The up line shows RPCM with different regularization norms on MNIST: from left to right, RPCM$_{\text{F}^2}$, RPCM$_{\ast}$, RPCM$_{\ell_1}$ and RPCM$_{\ell_1\hspace{-0.05cm}+\hspace{-0.01cm}\text{F}^2}$. Similarly, the down line shows RPCM with different regularization norms on NORB.}
\label{fig:parameteranalysis}
\end{center}
\vskip -0.1in
\end{figure*}

%\emph{Hopkins 155}\footnote{http://www.vision.jhu.edu/data/hopkins155/} consists of 155 video sequences, where 120 of the videos have two motions and 35 of the videos have three motions.

\begin{figure*}[ht]
\vskip -0.0in
\begin{center}
\centerline{\includegraphics[width=1.85\columnwidth]{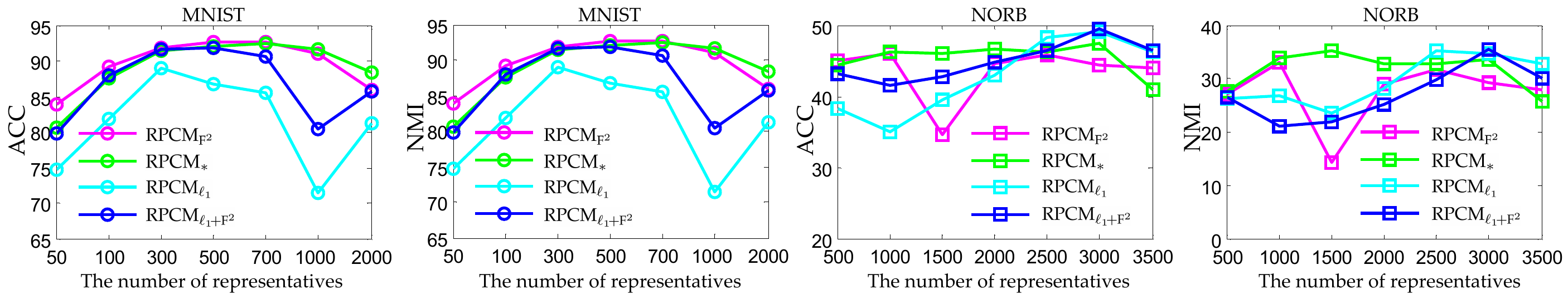}}
\vskip -0.1in
\caption{ACC $(\%)$ and NMI $(\%)$ with different number of representatives on MNIST and NORB.}
\label{fig:mnistdiffnumaccnmi}
\end{center}
\vskip -0.3in
\end{figure*}

\subsection{Baselines and Evaluations}
\textbf{Baselines.} To evaluate the performance of our LeaSC paradigm, we compare with nineteen state-of-the-art subspace clustering methods including three kinds classical coding (CCod), large-scale spectral clustering (LsSpC), and sampling-clustering-classification (S-C-C).

\emph{CCod} uses LSR \cite{Lucy2012}, SSC \cite{Elhamifar2013}, LRR \cite{Liugc2013}, ENSC \cite{Panagakis2014ENSC}, SSC-OPM\cite{Youc2016sssc}, and EnSC \cite{Youc2016ensc} to compute the codes of samples to build a similarity matrix, and apply NCuts \cite{Shij2000} to partition the matrix in small datasets. %All of them also are applied into middle and large datasets by choosing representatives as the dictionary and using LbSC \cite{Caid2015} to segment the data points.

\emph{LsSpC} fast constructs a similarity matrix by directly using the original data points. Nystr$\ddot{o}$m \cite{Chenwy2011} is to seek an approximate eigendecomposition of the similarity matrix, and Nystr$\ddot{o}$mO \cite{Chenwy2011} is further considered in the orthogonal eigenvectors. LbSC-K/-R \cite{Caid2015} choose some landmarks to build the similarity matrix by using $K$-means and random selections.

\emph{S-C-C} is to construct a \emph{sampling-clustering-classification} strategy. In particular, select+SSC (selSSC) and select+LRR (selLRR) \cite{Wangss2014} employ SSC and LRR to cluster the sampling data points, and train a simple linear classifier for segmenting the rest data points, while SLSR, SSSC and SLRR \cite{Pengx2016sc} cluster the sampling data points using LSR, SSC and LRR, and classify the rest data points using SRC or CRC \cite{Yang2012pr,Yang2013tnnls}.

\emph{LeaSC} is a machine learning paradigm that a parametric function is learned from the high-dimensional input space to their own subspaces for quickly dealing with large datasets. Based on different norms, in this paper we present five models (i.e., RPCM$_{\text{F}^2}$, RPCM$_{\ast}$, RPCM$_{\ell_1}$, and RPCM$_{\ell_1\hspace{-0.05cm}+\hspace{-0.01cm}\text{F}^2}$). In addition, many other related models (e.g., DAE\footnote{The codes are from https://github.com/kyunghyuncho/deepmat.} \cite{Vincent2010}, SSC-LOMP \cite{Lij2017sscl0}, RPCAec \cite{Sprechmann2015} and latent LRR (latLRR) \cite{Liu2011}) are also applied into our LeaSC paradigm.

\textbf{Evaluations.}
We measure the clustering results by using Normalized Mutual Information (NMI) and Clustering Accuracy (ACC) \cite{Caid2015}. They both change from 0 to 1. 0 shows total mismatching with the true subspace distribution, and 1 shows perfectly match. All experiments are repeated 10 times, and we report the mean and standard deviation of the final results.

 %between the produced clusters and the ground truth categories
\subsection{Parameter Settings}
In our RPCM model, there are one parametric function and two important parameters $\alpha$ and $\beta$ as $\overline{\alpha}=1$. We employ a three-layer neural network with 1,500 hidden units as the parametric function in all experiments. Its learning rate is set as $\zeta=1e^{-4}$, \emph{ReLU} is chosen as the activation function, and the number of training epochs is set to be less than $5$. We do experiments on MNIST and NORB. Fig. \ref{fig:parameteranalysis} shows the clustering results (i.e., ACC and NMI) with different $\alpha$ and $\beta$, which are selected from $\{1e^{-4},5e^{-4},1e^{-3},5e^{-3},1e^{-2},5e^{-2},0.1,1,10, 50, 100\}$.   Since there is no regularization term $R_2(\textbf{Z})$ in RPCM$_{\text{F}^2}$, RPCM$_{\ast}$, and RPCM$_{\ell_1}$, they are irrelevant to $\alpha$. From Fig. \ref{fig:parameteranalysis}, we observe that RPCM$_{\text{F}^2}$, RPCM$_{\ast}$, RPCM$_{\ell_1}$ and RPCM$_{\ell_1\hspace{-0.05cm}+\hspace{-0.01cm}\text{F}^2}$ achieve the best results at $\beta=1e^{-2}, 0.1, 0.1$, and $\alpha=100,\beta=10$ for MNIST and MNIST8M-1M, and $\beta=5e^{-2}, 1e^{-4}, 1$, and $\alpha=1,\beta=5e^{-2}$ for NORB, respectively. Moreover, they reach the good results at $\beta=5e^{-3}, 5e^{-3}, 1e^{-3}$, and $\alpha=1,\beta=0.5$ for YouTube8M-1M. In addition, the number of the selected representatives is also important to the clustering results. Using these parameters, Fig. \ref{fig:mnistdiffnumaccnmi} shows ACC and NMI with different number of representatives. This reveals that we select 500 and 3,000 representatives from MNIST and NORB for the best results, respectively.

%(i.e., Extended-YaleB and AR plotted in Fig. \ref{fig:imageExtended-YaleB-AR})
\begin{table}[t]
\vskip -0.0in
%\linespread{0.75}
\setlength{\tabcolsep}{3pt}
\renewcommand{\arraystretch}{1.2}
\caption{ACC $(\%)$ and NMI $(\%)$ on Extend-YaleB and AR.}
\vskip -0.15in
\label{tab:yalebarresults}
% \small
\begin{center}
\scalebox{1}{
\begin{tabular}{c|c|c|c|c|c}
\Xhline{1.2pt}
\multicolumn{2}{c|}{\multirow{2}*{Methods} } & \multicolumn{2}{c|}{Extended-YaleB}   & \multicolumn{2}{c}{AR}    \\
\cline{3-6}
 \multicolumn{2}{c|}{~}   & ACC   & NMI  & ACC   & NMI   \\
\hline
%\multicolumn{5}{c}{Learnable Subspace Clustering (LeaSC)}  \\
%\hline
\multirow{4}*{LeaSC}& RPCM$_{\ell_1\hspace{-0.05cm}+\hspace{-0.01cm}\text{F}^2}$  &73.5$\pm$0.72 & 75.4$\pm$0.86 & 81.3$\pm$1.28 & 90.9$\pm$0.56 \\
& RPCM$_{\ell_1}$                    & 72.2$\pm$0.87 & 73.0$\pm$0.61  & 80.8$\pm$1.05  & 90.5$\pm$0.44 \\
&RPCM$_{\ast}$                      & 59.6$\pm$1.25 & 61.1$\pm$0.73  & 62.3$\pm$0.95  & 70.2$\pm$0.59 \\
& RPCM$_{\text{F}^2}$                & 66.5$\pm$1.74 & 70.5$\pm$0.79  & 82.8$\pm$0.80  & 91.1$\pm$0.36  \\
%\hline
%\multicolumn{5}{c}{Classical Coding (CCod)}  \\
\hline
\multirow{4}*{CCod}&ENSC \cite{Panagakis2014ENSC}& 75.8$\pm$1.53 & 77.9$\pm$0.62  & 81.7$\pm$1.65 & 89.7$\pm$0.72 \\
&SSC \cite{Elhamifar2013}              & \textbf{76.5$\pm$1.22}& \textbf{78.4$\pm$0.35}  & 78.1$\pm$1.72  & 88.3$\pm$0.95  \\
&LRR \cite{Liugc2013}                  & 67.2$\pm$0.98 & 70.4$\pm$0.55  & 82.0$\pm$1.16  & \textbf{91.3$\pm$0.51}  \\
&LSR \cite{Lucy2012}                   & 67.0$\pm$1.17 & 70.7$\pm$0.51  & \textbf{83.1$\pm$1.21}  & \textbf{91.3$\pm$0.28}  \\
%\hline
%\multicolumn{5}{c}{Large-scale Spectral Clustering (LsSpC)} \\
\hline
\multirow{4}*{LsSpC}&LSR-R \cite{Caid2015} & 43.4$\pm$2.08 & 55.2$\pm$0.87  & 33.9$\pm$1.02  & 62.0$\pm$0.49  \\
&LSR-K \cite{Caid2015}                 & 43.4$\pm$1.21 & 54.5$\pm$0.63  & 34.4$\pm$0.72  & 62.5$\pm$0.67  \\
&Nystr$\ddot{o}$m \cite{Chenwy2011}    & 24.5$\pm$1.16 & 44.2$\pm$1.02  & 61.8$\pm$2.40  & 83.0$\pm$0.90  \\
&Nystr$\ddot{o}$mO \cite{Chenwy2011}   & 21.5$\pm$0.98 & 41.4$\pm$1.05  & 57.6$\pm$2.20  & 79.8$\pm$1.22  \\
\Xhline{1.2pt}
\end{tabular}}
\end{center}
\vskip -0.2in
\end{table}

\subsection{Clustering Results on Small Datasets}
In this subsection, we first validate that our RPCM models can efficiently approximate the clustering ACC and NMI of the CCod methods. We do experiments on small datasets (i.e., Extended-YaleB and AR), and select all data points as the representatives. For fair comparison, NCuts \cite{Shij2000} is used to obtain the final clustering results shown in Table \ref{tab:yalebarresults}. We observe that the performance of RPCM (i.e., RPCM$_{\text{F}^2}$, RPCM$_{\ast}$, RPCM$_{\ell_1}$ and RPCM$_{\ell_1\hspace{-0.05cm}+\hspace{-0.01cm}\text{F}^2}$) is comparable to CCod (i.e., LSR, LRR, SSC and ENSC) as RPCM trains a parametric function to effectively approximate to the ideal codes of CCod. Moreover, Table \ref{tab:timeonar} shows that the coding inference time of the parametric function learned by RPCM are 50 times faster than CCod at least. This reveals that RPCM can compute the codes in a rapid way although it takes some time to learn the parametric function. Additionally, Table \ref{tab:yalebarresults} also shows that RPCM is much better than LsSpC (e.g., LSR-R and LSR-K). Hence, the above observations drive us to apply RPCM into the LeaSC paradigm for large-scale datasets.

\begin{table}[t]
%\linespread{0.75}
\vskip -0.0in
\setlength{\tabcolsep}{3pt}
\renewcommand{\arraystretch}{1.2}
\caption{Inference time (second) comparison between LeaSC and CCod on Extend-YaleB and AR.}
\vskip -0.15in
\label{tab:timeonar}
\begin{center}
\scalebox{1}{
\begin{tabular}{c|c|c|c|c|c}
\Xhline{1.2pt}
\multicolumn{2}{c|}{\multirow{2}*{Methods} }  &\multicolumn{2}{c|}{Extend-YaleB} & \multicolumn{2}{c}{AR} \\
\cline{3-6}
 \multicolumn{2}{c|}{~} &  training   & coding   &training  & coding  \\
%\hline
%\multicolumn{5}{c}{Learnable Subspace Clustering (LeaSC)}  \\
\hline
\multirow{4}*{LeaSC}& RPCM$_{\ell_1\hspace{-0.05cm}+\hspace{-0.01cm}\text{F}^2}$  & 63  & \textbf{0.4}  & 79  & \textbf{0.2} \\
&RPCM$_{\ell_1}$                    & 58 & \textbf{0.4}  &73  & \textbf{0.2} \\
&RPCM$_{\ast}$                      & 43 & \textbf{0.4}  &64  & \textbf{0.2} \\
&RPCM$_{\text{F}^2}$                & 23 & \textbf{0.4}  &49  & \textbf{0.2} \\
%\hline
%\multicolumn{5}{c}{Classical Coding (CCod)}  \\
\hline
\multirow{4}*{CCod}&ENSC \cite{Panagakis2014ENSC}& 0  &61  & 0 & 77  \\
&SSC \cite{Elhamifar2013}    & 0  &56  & 0 & 72  \\
&LRR \cite{Liugc2013}        & 0  &41  & 0 & 63  \\
&LSR \cite{Lucy2012}         & 0  &20  & 0 & 48  \\
\Xhline{1.2pt}
\end{tabular}}
\end{center}
\vskip -0.2in
\end{table}

To compare our LeaSC paradigm with all nineteen subspace clustering methods, we do experiments on small subsets with 5,000 data points, which are randomly sampled from MNIST and NORB, respectively. In the subsets, we randomly selected 500 and 250 data points as the representatives for MNIST and NORB, respectively. We report all clustering results in Table \ref{tab:subsetresults}. We observe that RPCM$_{\text{F}^2}$ outperforms all the baseline methods on the subset in MNIST and it achieves at least $3.1\%$ (ACC) and $1.3\%$ (NMI) improvement. On the subset in NORB,  RPCM$_{\text{F}^2}$ is also better than the most of the baseline methods. However, RPCM$_{\text{F}^2}$ is less than SSC and LSR, which show the super clustering ability of the CCod methods on small datasets. Thus, this observation also drives us to extend the CCod methods into large scale datasets. Note that RPCM$_{\ell_1\hspace{-0.05cm}+\hspace{-0.01cm}\text{F}^2}$ and RPCM$_{\ell_1}$ are less than RPCM$_{\text{F}^2}$ and RPCM$_{\ast}$ on the subset in NORB. A plausible reason is that the number of representatives does not satisfy the over-complete assumption for sparse coding models. Thus, we select 3,000 representatives in the whole NORB dataset. %in the next subsection.
%The final results were reported as the average and variance of 10 times repetition.
\begin{table}[t]
%\linespread{0.75}
\vskip -0.0in
\setlength{\tabcolsep}{3pt}
\renewcommand{\arraystretch}{1.2}
\caption{ACC $(\%)$ and NMI $(\%)$ on small subsets selected 5000 samples from MNIST and NORB. }
\vskip -0.15in
\label{tab:subsetresults}
\begin{center}
%\footnotesize{
\scalebox{0.95}{
\begin{tabular}{c|c|c|c|c|c}
\Xhline{1.2pt}
\multicolumn{2}{c|}{\multirow{2}*{Methods} }  & \multicolumn{2}{c|}{MNIST(5,000)} & \multicolumn{2}{c}{NORB(5,000)} \\
\cline{3-6}
 \multicolumn{2}{c|}{~}                        & ACC   & NMI        & ACC   & NMI   \\
%\hline
%\multicolumn{5}{c}{Learnable Subspace Clustering (LeaSC)}  \\
\hline
\multirow{8}*{LeaSC} & RPCM$_{\ell_1\hspace{-0.05cm}+\hspace{-0.01cm}\text{F}^2}$  &94.9$\pm$0.28&88.3$\pm$0.50&40.4$\pm$2.28&21.9$\pm$1.49\\
&RPCM$_{\ell_1}$                  &88.3$\pm$4.47&81.3$\pm$2.50 &41.9$\pm$3.16&23.8$\pm$2.50\\
&RPCM$_{\ast}$                    &95.9$\pm$0.37&90.2$\pm$0.67 &44.4$\pm$1.21&26.7$\pm$3.69  \\
&RPCM$_{\text{F}^2}$             &\textbf{96.2$\pm$0.53}&\textbf{91.0$\pm$0.89} &47.5$\pm$0.86&35.0$\pm$1.28 \\
\cline{2-6}
&SSC-LOMP \cite{Lij2017sscl0}    &93.1$\pm$0.92&89.7$\pm$1.34 &32.7$\pm$3.03&9.38$\pm$3.80 \\
&DAE \cite{Vincent2010}          &91.2$\pm$3.21&84.3$\pm$1.92 &33.1$\pm$4.36&18.9$\pm$2.93 \\
&latLRR \cite{Liu2011}           &92.3$\pm$4.39&85.6$\pm$1.86 &35.2$\pm$3.39&20.0$\pm$2.84\\
&RPCAec \cite{Sprechmann2015}    &90.8$\pm$3.91&83.2$\pm$2.28 &37.0$\pm$4.44&21.8$\pm$7.08 \\
%\hline
%\multicolumn{5}{c}{Classical Coding (CCod)}  \\
\hline
\multirow{6}*{CCod} &EnSC \cite{Youc2016ensc}     &90.5$\pm$7.00&89.0$\pm$2.66 &39.4$\pm$7.78&17.8$\pm$8.26 \\
&ENSC \cite{Panagakis2014ENSC}&90.2$\pm$6.10&86.5$\pm$2.33 &41.9$\pm$1.81&23.5$\pm$1.23 \\
&SSC-OPM\cite{Youc2016sssc}   &89.8$\pm$1.99&84.1$\pm$1.79 &28.5$\pm$3.16&6.30$\pm$2.54\\
&SSC \cite{Elhamifar2013}     &83.8$\pm$4.27&88.7$\pm$2.13 &\textbf{49.7$\pm$5.87}&\textbf{37.7$\pm$9.11}    \\
&LRR \cite{Liugc2013}         &80.7$\pm$1.55&81.2$\pm$1.74 &43.9$\pm$1.96&26.3$\pm$0.91    \\
&LSR \cite{Lucy2012}          &80.7$\pm$0.98&75.6$\pm$1.63 &48.9$\pm$1.63&28.8$\pm$1.40 \\
%\hline
%\multicolumn{5}{c}{Sampling-Clustering-Classification (S-C-C)}\\
\hline
\multirow{5}*{S-C-C} &SSSC \cite{Pengx2016sc}            &78.9$\pm$3.75&77.9$\pm$2.20 &39.5$\pm$1.59&17.8$\pm$2.55    \\
&SLSR \cite{Pengx2016sc}            &78.6$\pm$7.12&76.4$\pm$3.82 &38.1$\pm$0.71&16.0$\pm$0.60   \\
&SLRR \cite{Pengx2016sc}            &81.7$\pm$5.87&79.3$\pm$2.96 &39.9$\pm$1.24&18.3$\pm$1.04  \\
&selSSC \cite{Wangss2014}           &68.8$\pm$5.89&70.0$\pm$2.98 &36.0$\pm$3.35&16.5$\pm$6.75\\
&selLRR \cite{Wangss2014}           &68.3$\pm$5.75&64.4$\pm$2.29 &39.7$\pm$2.08&18.2$\pm$2.62\\
%\hline
%\multicolumn{5}{c}{Large-scale Spectral Clustering (LsSpC)} \\
\hline
\multirow{4}*{LsSpC} &LbSC-K \cite{Caid2015}             &90.6$\pm$4.03&83.9$\pm$1.66 &41.3$\pm$3.19&25.3$\pm$3.03 \\
&LbSC-R \cite{Caid2015}             &85.0$\pm$5.95&78.6$\pm$2.28 &39.0$\pm$4.43&21.4$\pm$5.37  \\
&Nystr$\ddot{o}$mO \cite{Chenwy2011}&74.8$\pm$6.68&67.4$\pm$2.88 &37.7$\pm$3.41&19.0$\pm$0.16 \\
&Nystr$\ddot{o}$m \cite{Chenwy2011} &72.2$\pm$8.31&66.1$\pm$4.31 &37.1$\pm$7.21&17.8$\pm$1.63 \\
\Xhline{1.2pt}
\end{tabular}}
\end{center}
\vskip -0.2in
\end{table}

%The final results were reported as the average and variance of 10 times repetition.
\begin{table*}[t]
%\linespread{0.75} %RPCM$_{\ell_1\hspace{-0.05cm}+\hspace{-0.05cm}\ast}$       &  $\pm$  &  $\pm$  &  $\pm$  &  $\pm$ &  $\pm$  &  $\pm$  &  $\pm$  &  $\pm$  \\
\vskip -0.0in
\setlength{\tabcolsep}{5pt}
\renewcommand{\arraystretch}{1.2}
\caption{ACC $(\%)$ and NMI $(\%)$ on large datasets. We randomly selected 500, 3000, 500 and 500 samples as representatives from MNIST, NORB, MNIST8M-1M and YouTube8M-1M. }
\vskip -0.15in
\label{tab:largedataresults}
\begin{center}
%\footnotesize{
\scalebox{1}{
\begin{tabular}{c|c|c|c|c|c|c|c|c|c}
\Xhline{1.2pt}
\multicolumn{2}{c|}{\multirow{2}*{Methods} } & \multicolumn{2}{c|}{MNIST} &\multicolumn{2}{c|}{NORB} & \multicolumn{2}{c|}{MNIST8M-1M}  & \multicolumn{2}{c}{YouTube8M-1M} \\
\cline{3-10}
 \multicolumn{2}{c|}{~}         & ACC   & NMI        & ACC   & NMI   & ACC   & NMI   & ACC   & NMI      \\
%\hline
%\multicolumn{9}{c}{Learnable Subspace Clustering (LeaSC)}  \\
\hline
\multirow{8}*{LeaSC}&RPCM$_{\ell_1\hspace{-0.05cm}+\hspace{-0.01cm}\text{F}^2}$ &96.9$\pm$0.19&91.8$\pm$0.39 &\textbf{49.5$\pm$5.68}&\textbf{35.6$\pm$5.77} &95.6$\pm$0.16&89.3$\pm$0.29 &23.1$\pm$0.21&38.1$\pm$0.26\\
&RPCM$_{\ell_1}$              &94.6$\pm$2.45&88.7$\pm$2.36 &49.1$\pm$4.00&34.7$\pm$6.98 &92.2$\pm$2.18&83.8$\pm$2.99&25.8$\pm$0.66&43.1$\pm$0.95\\
&RPCM$_{\ast}$          &97.0$\pm$0.10&92.1$\pm$0.18 &47.5$\pm$4.62&33.7$\pm$4.49 &95.9$\pm$0.10&89.9$\pm$0.16&25.8$\pm$0.13&42.8$\pm$0.02\\
&RPCM$_{\text{F}^2}$    &\textbf{97.2$\pm$0.07}&\textbf{92.6$\pm$0.16} &44.5$\pm$4.77&29.3$\pm$9.36 &\textbf{96.1$\pm$0.15}&\textbf{90.2$\pm$0.27}&\textbf{26.0$\pm$0.31}&\textbf{43.5$\pm$0.03}   \\
\cline{2-10}
&SSC-LOMP \cite{Lij2017sscl0}    &91.5$\pm$5.58&84.8$\pm$2.40 &43.2$\pm$1.54&24.9$\pm$1.59 &91.6$\pm$1.28&82.6$\pm$1.25&23.7$\pm$0.58&40.1$\pm$0.02\\
&DAE \cite{Vincent2010}          &88.0$\pm$5.90&82.4$\pm$2.80 &36.8$\pm$3.79&17.1$\pm$4.49 &71.9$\pm$3.38&65.9$\pm$1.39&24.1$\pm$0.69&40.7$\pm$0.50\\
&latLRR \cite{Liu2011}           &88.8$\pm$5.85&83.1$\pm$2.47 &39.0$\pm$0.45&21.6$\pm$0.99 &93.4$\pm$0.50&84.7$\pm$0.81&23.8$\pm$0.49&39.9$\pm$0.33\\
&RPCAec \cite{Sprechmann2015}    &71.0$\pm$5.44&64.6$\pm$4.06 &45.3$\pm$0.97&31.2$\pm$2.53 &92.6$\pm$4.66&86.4$\pm$2.79&20.8$\pm$0.42&32.0$\pm$0.32\\
%\hline
%\multicolumn{9}{c}{Sampling-Clustering-Classification (S-C-C)}\\
\hline
\multirow{5}*{S-C-C}&SSSC \cite{Pengx2016sc}&78.9$\pm$4.88&76.4$\pm$3.25 &39.3$\pm$2.00&15.1$\pm$3.06&78.4$\pm$6.54&72.5$\pm$4.07 &22.8$\pm$0.51&38.2$\pm$0.55\\
&SLSR \cite{Pengx2016sc}    &75.4$\pm$5.60&70.6$\pm$2.91 &39.9$\pm$0.31&20.6$\pm$0.15 &71.2$\pm$5.17&65.1$\pm$3.18&18.5$\pm$0.65&32.4$\pm$0.68\\
&SLRR \cite{Pengx2016sc}    &79.7$\pm$0.68&76.5$\pm$1.44 &40.1$\pm$0.74&21.6$\pm$0.81 &78.1$\pm$5.69&73.0$\pm$3.50&20.1$\pm$0.42&37.6$\pm$0.56\\
&selSSC \cite{Wangss2014}   &74.7$\pm$5.58&69.9$\pm$2.93 &36.0$\pm$3.36&18.9$\pm$6.75 &73.3$\pm$5.82&68.5$\pm$3.51&18.8$\pm$0.72&32.4$\pm$0.78 \\
&selLRR \cite{Wangss2014}   &75.4$\pm$4.31&69.2$\pm$2.27 &40.6$\pm$0.18&21.5$\pm$0.21 &74.5$\pm$5.31&68.8$\pm$3.29&19.6$\pm$0.56&33.7$\pm$0.62\\
%\hline
%\multicolumn{9}{c}{Large-scale Spectral Clustering (LsSpC)} \\
\hline
\multirow{4}*{LsSpC}&LbSC-K \cite{Caid2015} &85.9$\pm$0.34&80.2$\pm$0.45&45.5$\pm$1.35&30.9$\pm$0.52&88.5$\pm$0.73&79.6$\pm$0.71 &24.9$\pm$0.59&42.1$\pm$0.50\\
&LbSC-R \cite{Caid2015}  &81.2$\pm$1.06&73.9$\pm$1.11 &43.3$\pm$2.29&27.6$\pm$2.33 &79.3$\pm$3.89&70.6$\pm$1.15 &13.7$\pm$0.78&24.4$\pm$1.35\\
&Nystr$\ddot{o}$mO \cite{Chenwy2011}&71.0$\pm$7.43&63.9$\pm$3.65 &37.3$\pm$0.88&18.9$\pm$0.25 &62.3$\pm$3.19&50.6$\pm$1.83&12.2$\pm$1.32&22.4$\pm$1.83\\
&Nystr$\ddot{o}$m \cite{Chenwy2011} &70.8$\pm$7.56&64.2$\pm$3.58 &37.1$\pm$0.84&18.7$\pm$0.59 &61.1$\pm$2.64&50.6$\pm$1.88&11.6$\pm$1.42&22.2$\pm$1.95\\
\Xhline{1.2pt}
\end{tabular}}
\end{center}
\vskip -0.2in
\end{table*}

\subsection{Clustering Results on Large Datasets}
Since it is well-known that the CCod methods (e.g., LRR and SSC) are difficult (or impossible) to compute the codes in large datasets with over ten
thousand data points, we propose the LeaSC paradigm to quickly calculate the ideal codes for large-scale subspace clustering. Since the DS3 model \cite{Elhamifar2016subset} is solved by the sparse coding method, it costs much time to choose the representative. For saving the chosen time, we randomly select 500, 3000, 500 and 500 samples as the representatives from MNIST, NORB, MNIST8M-1M and YouTube8M-1M respectively. We report the final clustering results in Table \ref{tab:largedataresults} and have the following observations:

First, we can see that our RPCM models outperform all baseline methods. RPCM has much better clustering results than the S-C-C methods (e.g., SLSR, SLRR, SSSC, selSSC, and selLRR). Particularly, RPCM$_{\text{F}^2}$ achieves at least $18.3\%$ (ACC) and $12.1\%$ (NMI) improvement on MNIST, RPCM$_{\ell_1\hspace{-0.05cm}+\hspace{-0.01cm}\text{F}^2}$ reaches $8.9\%$ (ACC) and $14.0\%$ (NMI) improvement on NORB, and RPCM$_{\text{F}^2}$ also gets at least $17.7\%$ (ACC) and $17.2\%$ (NMI) improvement on MNIST8M-1M, and obtains at least $3.2\%$ (ACC) and $5.3\%$ (NMI) improvement on YouTube8M-1M. The basic reason is that the simple linear classifier in the S-C-C methods does not have the power to handle the complex examples. In contrast, RPCM can fast capture the excellent codes of the complex examples by training the neural networks, and then performs LSC-K.

%\begin{figure}[t]
%\vskip -0.0in
%\begin{center}
%\centerline{\includegraphics[width=0.98\columnwidth]{youtube-8m-samples.pdf}}
%\vskip -0.1in
%\caption{Four videos in YouTube-8M-1M: from up to down, Boat, Wrestling, Smartphone, and Racing.}
%\label{fig:videosYouTube-8M-1M}
%\end{center}
%\vskip -0.3in
%\end{figure}

%\textbf{Fast clustering time.}

\begin{table}[t]
%\linespread{0.75}
\vskip -0.0in
\setlength{\tabcolsep}{2.6pt}
\caption{All clustering time (second) compared RPCM with S-C-C on MNIST, NORB, MNIST8M-1M and YouTube8M-1M.}
\vskip -0.15in
\label{tab:timeonmnist}
\begin{center}
\scalebox{1}{
\begin{tabular}{c|c|c|c|c}
\Xhline{1.2pt}
Methods & MNIST & NORB  & MNIST8M-1M  &YouTube8M-1M \\
\hline
RPCM$_{\ell_1\hspace{-0.05cm}+\hspace{-0.01cm}\text{F}^2}$  &  80  &  358  & 436  & 398\\
RPCM$_{\ell_1}$              &  83  &  437  &  460  & 426\\
RPCM$_{\ast}$                &  55  &   710 &  412  & 383  \\
RPCM$_{\text{F}^2}$          &  \textbf{35}  &\textbf{63}  & \textbf{396}  & \textbf{368}  \\
\hline
SLSR \cite{Pengx2016sc}      &  274  &1365  &  3623  &  3161  \\
SLRR \cite{Pengx2016sc}      &  291  &2376  &  3798   &  3252  \\
SSSC \cite{Pengx2016sc}      &  285  &2353  &   3755  &   3237 \\
selSSC \cite{Wangss2014}     &  469  &3629  &   $>$24h &   $>$24h \\
selLRR \cite{Wangss2014}     &  1569 &4489  &   $>$24h &   $>$24h  \\
\Xhline{1.2pt}
\end{tabular}}
\end{center}
\vskip -0.2in
\end{table}
% 57.65, 54.32 33.28 13.32 one time

Second, since RPCM can learn the excellent codes of the original samples, RPCM still has higher ACC and NMI than the LsSpC methods without coding models (e.g., LSR-R, LSR-K, Nystr$\ddot{o}$m, and Nystr$\ddot{o}$mO). Specifically, RPCM$_{\text{F}^2}$ achieves at least $11.3\%$ (ACC) and $10.4\%$ (NMI) improvement on MNIST, and RPCM$_{\ell_1\hspace{-0.05cm}+\hspace{-0.01cm}\text{F}^2}$ reaches $4.0\%$ (ACC) and $4.7\%$ (NMI) improvement on NORB. RPCM$_{\text{F}^2}$ also gets better $7.6\%$ (ACC) and $10.6\%$ (NMI) on MNIST8M-1M, and obtains better $1.1\%$ (ACC) and $1.4\%$ (NMI) on YouTube8M-1M.

Third, RPCM$_{\text{F}^2}$ is better than RPCM$_{\ast}$, RPCM$_{\ell_1}$, RPCM$_{\ell_1\hspace{-0.05cm}+\hspace{-0.01cm}\text{F}^2}$ on the large datasets except for NORB. An important reason is that the experiment results are based on the 'good' scattering or deep convolutional features on the large datasets, while we use the raw pixels in NORB. In addition, our RPCM models are better than the direct encoding models (e.g., DAE \cite{Vincent2010}, SSC-LOMP \cite{Lij2017sscl0}, RPCAec \cite{Sprechmann2015} and latent LRR (latLRR) \cite{Liu2011}) in many cases because the RPCM models can learn robust representations by separating the sparse noise. Although latLRR also can separate the sparse noise, it learn a linear mapping to calculate the codes from the original samples. Unfortunately, the linear mapping is difficult to capture complex and nonlinear information.

Fourth, the overall clustering time are reported in Table \ref{tab:timeonmnist}, which includes the training, coding, and clustering times. We can see that RPCM is faster than S-C-C because RPCM can quickly infer the representations by only calculating the trained neural networks. We also observe that RPCM is at least three times faster than SLSR, SLRR, SSSC, selLRR, and selSSC. In addition, although RPCM is slower than LsSpC, it perform better than LsSpC. In fact, when the number of data points increases, its time complexity is comparable to LsSpC since the training time can be ignored.

\begin{figure}[t]
\vskip -0.0in
\begin{center}
\centerline{\includegraphics[width=0.85\columnwidth]{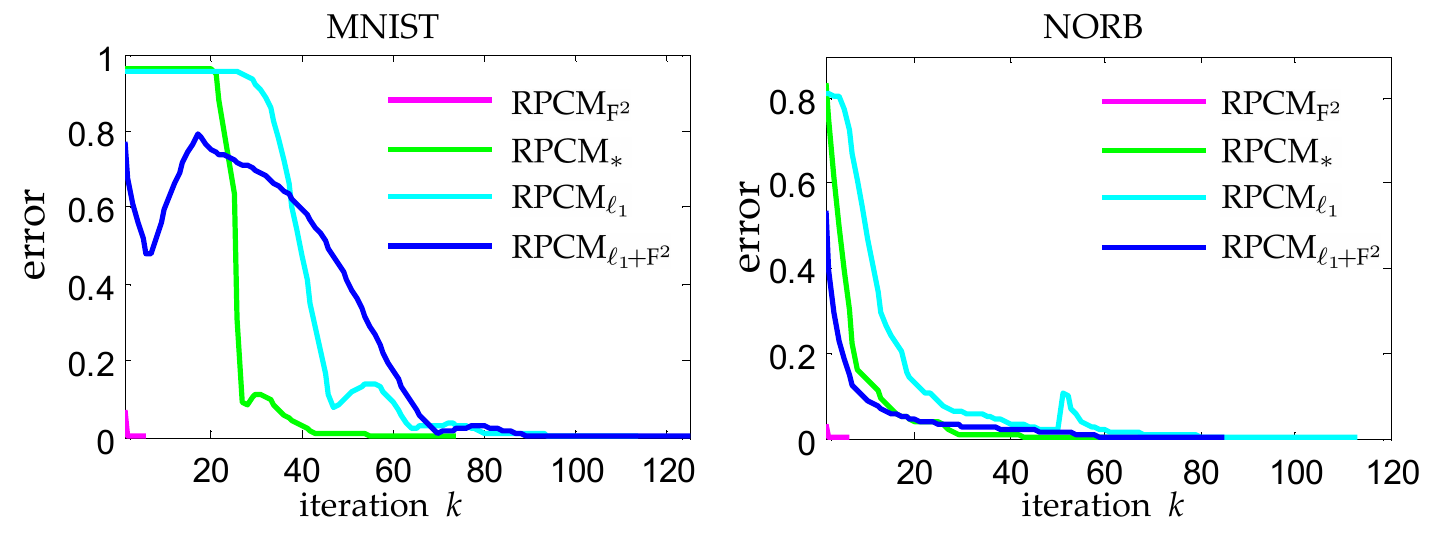}}
\vskip -0.1in
\caption{Convergence of RPCM with different regularization norms on MNIST (left) and NORB (right).}
\label{fig:convergence}
\end{center}
\vskip -0.2in
\end{figure}

Fifth, we verify the convergence of our Algorithm \ref{alg:RPCM}, and the error curves are plotted in Fig. \ref{fig:convergence}. We only train the neural network in RPCM$_{\ell_2}$ five epochs as $\textbf{Z}$ has a closed solution.

%In addition, RCC is better and faster than DAE and PSD in the FCod methods. Besides, RCC is also better than LSR as RCC trains a non-linear function.

%We also proposed an algorithm by alternating ADMM and GD to minimize the RPCM model.

\section{Conclusion}
\label{sec:cons}
To effectively handle with million-scale datasets in subspace clustering problems, we presented an efficient LeaSC paradigm. First, a representative set was sampled from the large-scale dataset. Second, using the small representative set, we proposed an RPCM model to train a parametric function from the high-dimensional subspaces to the low-dimensional subspaces. Thirdly, the trained parametric function is used to quickly compute the contractive low-dimensional representations of all data points, and the large-scale spectral clustering methods can be used to cluster the representations for the final clustering results. Besides, we provided a contractive analysis of the parametric function to show its effectiveness for subspace clustering. Experimental results verified that our LeaSC successfully deal with million-scale datasets.

\appendices

\section{Proof of Proposition 1}
\emph{Proof:} Before proving the results, we firstly give some notations. We suppose that a data matrix $\textbf{Y}\in{\rm I\!R}^{d\times m}$ is sampled from a union of subspaces with $\widehat{n}$ bases.
\begin{itemize}
  \item \textbf{Dividing $\textbf{Y}$.} Depending on the bases, $\textbf{Y}$ is divided into $\widehat{n}$ sub-matrices $\{\textbf{Y}_i\in{\rm I\!R}^{d\times m_i}\}_{i=1}^{\widehat{n}}$, where each data point in $\textbf{Y}_i$ can be regarded as the $i$th basis and $m_i$ is the number of data points in $\textbf{Y}_i$ ($\sum_{i=1}^{\widehat{n}}m_i=m$).
  \item \textbf{Dividing $\textbf{X}$.} Similarly, when we randomly choose a representative matrix $\textbf{X}\in{\rm I\!R}^{d\times n}$ from $\textbf{Y}$, it is divided into $\widehat{n}$ sub-matrices $\{\textbf{X}_i\in{\rm I\!R}^{d\times n_i}\}_{i=1}^{\widehat{n}}$, where $\textbf{X}_i\subset\textbf{Y}_i$ and $n_i$ is the number of data points in $\textbf{X}_i$ ($\sum_{i=1}^{\widehat{n}}n_i=n$).
  \item \textbf{An event.} In order to ensure that $\textbf{X}$ includes all the bases, that is, each $\textbf{X}_i$ \emph{includes at least one data point in} $\textbf{Y}_i$ (i.e., $1 \leq n_i\leq m_i$), this is denoted by an event $(n_1,\cdots,n_{\widehat{n}})$, which satisfies $\sum_{i=1}^{\widehat{n}}n_i=n $ and $\{1 \leq n_i\leq m_i\}_{i=1}^{\widehat{n}}$.
  \item \textbf{A set $\mathbf{S}$ of all possible events.} Following the event, all possible events are denoted by $\mathbf{S}=\{(n_1,\cdots,n_{\widehat{n}})|\sum_{i=1}^{\widehat{n}}n_i=n \& \{1 \leq n_i\leq m_i\}_{i=1}^{\widehat{n}}\}$.
\end{itemize}

Now, we prove a probability $P$ to guarantee that each base has at least one select for effective subspace clustering when we random select $\textbf{X}$ from $\textbf{Y}$. Based on the event, this probability is a type of ratio where we compare how many times all possible events can occur compared to all possible selects, that is
\begin{align}
\label{eq:probselects1}
P=\frac{\text{The number of all possible events}}{\text{The number of all possible selects}},
\end{align}
where
\begin{itemize}
  \item \emph{The number of all possible selects.} The number of selecting $n$ data points $\textbf{X}$ from $m$ data points $\textbf{Y}$ is the $n$-combination from a set including $m$ elements, $\mathrm{C}_m^n$.
  \item \emph{The number of all possible events.} In each event $(n_1,\cdots,n_{\widehat{n}})$, the selected number of the event is $\prod_{i=1}^{\widehat{n}}\mathrm{C}_{m_i}^{n_i}$. When considering all possible events, the selected number is a sum of the selected numbers of all events, $\sum_{(n_1,\cdots,n_{\widehat{n}})\in\mathbf{S}}\left(\prod_{i=1}^{\widehat{n}}\mathrm{C}_{m_i}^{n_i}\right)$.
\end{itemize}
Thus, the probability $P$ is rewritten as the Eq. \eqref{eq:probselects}. $\blacksquare$

\textbf{Remark A1.} Depending on the choosing number $n$, there are three cases to show the probability $P$.
\begin{enumerate}
  \item When $0<n<\widehat{n}$, $\mathbf{S}=\emptyset$, that is, we cannot select $\textbf{X}$ including all the bases. Thus, its probability is zero, $P=\frac{0}{\mathrm{C}_m^n}=0$;
  \item When $\widehat{n}\leq n\leq \max_{i}\{m-m_i\}$, it has inequalities $0<\sum_{(n_1,\cdots,n_{\widehat{n}})\in\mathbf{S}}\left(\prod_{i=1}^{\widehat{n}}\mathrm{C}_{m_i}^{n_i}\right)<\mathrm{C}_m^n$. Thus, the probability is $0<P<1$;
  \item When $\max_{i}\{m-m_i\}<n\leq m$, no matter how we choose, it gets $\textbf{X}$, which always includes all the bases, that is $\sum_{(n_1,\cdots,n_{\widehat{n}})\in\mathbf{S}}\left(\prod_{i=1}^{\widehat{n}}\mathrm{C}_{m_i}^{n_i}\right)=\mathrm{C}_m^n$. Thus, we always have $P=1$.
\end{enumerate}

\section{Proof of Theorem 1}

\emph{Proof:} By using the first-order Taylor series with the representative $\textbf{x}_{ij}\in\textbf{X}_i\subset\textbf{X}$ $(1\leq j\leq N_i, 1\leq i\leq s)$, for any
$\textbf{y}\in\textbf{Y}_{ij}\subset\textbf{Y}_i\subset\textbf{Y}\subset\mathcal{N}_{\textbf{X}}^\rho$, we obtain
\begin{align}
\label{eq:Htaylorsupp}
f(\textbf{y};\theta)=&f(\textbf{x}_{ij};\theta)+\frac{\partial f(\textbf{x}_{ij};\theta)}{\partial \textbf{x}_{ij}}\left(\textbf{y}-\textbf{x}_{ij}\right) \nonumber \\
&+O\left(\textbf{y}-\textbf{x}_{ij}\right).
\end{align}

Since the distance between $\textbf{y}$ and $\textbf{x}_{ij}$ is less than $\rho$, it holds that
\begin{align}
\label{eq:contractionresultsupp1}
\|f(\textbf{x}_{ij};\theta)-f(\textbf{y};\theta)\|\leq \|\frac{\partial f(\textbf{x}_{ij};\theta)}{\partial \textbf{x}_{ij}}\|_{\textbf{F}}\rho+O(\rho).
\end{align}

If $\|\frac{\partial f(\textbf{x}_{ij};\theta)}{\partial \textbf{x}_{ij}}\|_{\textbf{F}}\leq \rho$, then the Eq. \eqref{eq:contractionresultsupp1} is rewritten as
\begin{align}
\label{eq:contractionresultsupp2}
\|f(\textbf{x}_{ij};\theta)-f(\textbf{y};\theta)\| &\leq \rho^2+O(\rho) \nonumber \\
&\leq O(\rho).
\end{align}

By merging Eq. \eqref{eq:contractionresultsupp1} with Eq. \eqref{eq:contractionresultsupp2}, it is easy to check the condition. $\blacksquare$


% or


\ifCLASSOPTIONcaptionsoff
  \newpage
\fi

%\bibliographystyle{ieee}
\bibliographystyle{IEEEtran}
\bibliography{curRefs}

\end{document}